\documentclass[12pt]{article}

\usepackage{sbc-template}
\usepackage{graphicx,url}
\usepackage{verbatim}
\usepackage{listings}
\usepackage{indentfirst}
\usepackage[T5]{fontenc}
\usepackage[utf8]{inputenc}
\usepackage{amsmath}
\usepackage{tabularx}
\usepackage{longtable}
\usepackage{caption2}

\title{Vintern-1B: An Efficient Multimodal Large Language Model for Vietnamese}

\author{Khang T. Doan\inst{1}, Bao G. Huynh\inst{2} , Dung T. Hoang\inst{3}, Thuc D. Pham\inst{4}, \\ Nhat H. Pham\inst{5}, Quan T.M. Nguyen\inst{6}, Bang Q. Vo\inst{7}, Suong N. Hoang\inst{8}}

\address{dtkhangbk@gmail.com \inst{2}huynhgiabaoa2@gmail.com \\ \inst{3}htdung167@gmail.com \inst{4}phamdinhthuc020100@gmail.com \\ \inst{5}phamhuynhnhat99@gmail.com \inst{6}minhquan240600@gmail.com \\
\inst{7}bavo.imp@gmail.com
\inst{8}suong.nhoang@gmail.com
}

\begin{document} 
\maketitle

\begin{abstract}
    
   In this report, we introduce Vintern-1B (Vietnamese-InternVL2-1B), a reliable 1-billion-parameters multimodal large language model (MLLM) for Vietnamese language tasks. By integrating the Qwen2-0.5B-Instruct \cite{yang2024qwen2} language model with the InternViT-300M-448px \cite{chen2024internvl} visual model, Vintern-1B is optimized for a range of applications, including optical character recognition, document information extraction, and general visual question-answering in Vietnamese context. The model is fine-tuned on an extensive dataset of over 3 million image-question-answer pairs, achieving robust performance and reliable results across multiple Vietnamese language benchmarks like OpenViVQA \cite{nguyen2023openvivqa}. Vintern-1B is small enough to fit into various on-device applications easily. Additionally, we have open-sourced several Vietnamese vision question answering (VQA) datasets for text, documents, diagrams, and more created with Gemini 1.5 Flash \cite{reid2024gemini}. Our models are available at: \url{https://huggingface.co/5CD-AI/Vintern-1B-v2}.

\end{abstract}

\section{Introduction}

    The emergence of large language models (LLMs) has significantly transformed the field of natural language processing, demonstrating exceptional capabilities in tasks requiring complex reasoning and deep linguistic comprehension. Building on the success of LLMs, researchers have begun to explore multimodal large language models (MLLMs) \cite{chen2024internvl} \cite{reid2024gemini} \cite{bai2023qwenvlversatilevisionlanguagemodel} \cite{liu2024improvedllava1.6} \cite{li2024llavanext} \cite{achiam2023gpt4}, which integrate visual data with textual information. These models have proven highly effective in tasks such as image captioning, visual question answering, and multimodal machine translation, where understanding the interaction between language and vision are crucial.
    
    Despite global progress in developing MLLMs, the advancement of Vietnamese MLLMs has been hampered by the limited availability of high-quality multimodal datasets. While models like Qwen-VL \cite{bai2023qwenvlversatilevisionlanguagemodel}, MiniCPM-Llama3-V-2.5 \cite{yao2024minicpm}, MoE-LLaVA-Qwen \cite{lin2024moellava}, and InternVL \cite{chen2024internvl} have shown promise in processing Vietnamese text, there remains a gap in resources and models fully supporting Vietnamese multimodal tasks.
    
    In Vietnam, initiatives such as the V-Vistral \cite{ViVLMVistralVision2024} model, the Vista dataset \cite{ViVLMVista2024}, and the development of LaVy \cite{tran2024lavy} have made significant contributions to the field. However, these models and resources currently lack support for Vietnamese OCR-related tasks and document processing, which are critical for a wide range of practical applications.
    
    To address these limitations, we carefully created datasets with a focus on text and document processing, as well as additional datasets featuring Vietnamese localization images. We then fine-tuned the InternVL2-1B model, which has achieved impressive results in text processing with our datasets. We provide a detailed description of Vintern-1B’s architecture, data creation, training procedures, and how to benchmark our MLLMs.
    
    Through these contributions, we aim to advance the field of Vietnamese MLLMs, equipping researchers and practitioners with the tools and resources necessary to explore and innovate at the intersection of language and vision within the Vietnamese context.

\section{Related work} \label{sec:firstpage}


\subsection{Multimodal Large Language Model}



Closed-source MLLMs, such as GPT-4o \cite{2023GPT4VisionSC} and Gemini Pro Vision \cite{team2023gemini}, have significantly advanced the integration of visual inputs, enabling sophisticated handling of text, images, and audio. Concurrently, the past year has witnessed a surge in open-source MLLMs, where researchers have been striving to achieve unified reasoning across modalities by leveraging the advancements in LLMs. A notable contribution in this space is the InternVL series \cite{bai2023qwenvlversatilevisionlanguagemodel} \cite{chen2024internvl}, which has showcased impressive results by aligning visual inputs with LLMs through visual instruction tuning.

InternVL2-1B \cite{chen2024internvl}, for instance, employs the InternViT-300M-448px, a distilled vision foundation model with 300M parameters derived from the robust vision encoder InternViT-6B-448px, which possesses continuous learning capabilities. This enhancement significantly boosts its visual understanding, enabling seamless transfer and reuse across different LLMs. The language model Qwen2-0.5B-Instruct \cite{yang2024qwen2}, which has been integrated into this framework, demonstrates competitive multilingual capabilities, proficient in approximately 30 languages, particularly excelling in Vietnamese, promising strong support for MLLMs designed for the Vietnamese language.

\subsection{Vietnamese MLLMs}
In the Vietnamese language context, recent models like V-Vistral \cite{ViVLMVistralVision2024} and LaVy \cite{tran2024lavy} have made notable progress in advancing MLLMs. However, these models still struggle with understanding documents, charts, infographics, and recognizing scene texts specific to Vietnamese. While large Vietnamese MLLM datasets like Vistra \cite{ViVLMVista2024}, which was translated from English to Vietnamese using Gemini Pro, have enhanced the naturalness of question-answering, the overall model performance remains suboptimal. This disparity is mainly due to the lack of high-quality, purely Vietnamese image content. The situation is further complicated by the absence of diverse datasets that support various tasks such as OCR and information extraction, which are essential for improving the capabilities of MLLMs in Vietnamese.

\section{Vintern-1B}
\subsection{Overall Architecture}

\begin{figure}[t!]
\centering
\includegraphics[width=1.0\columnwidth]{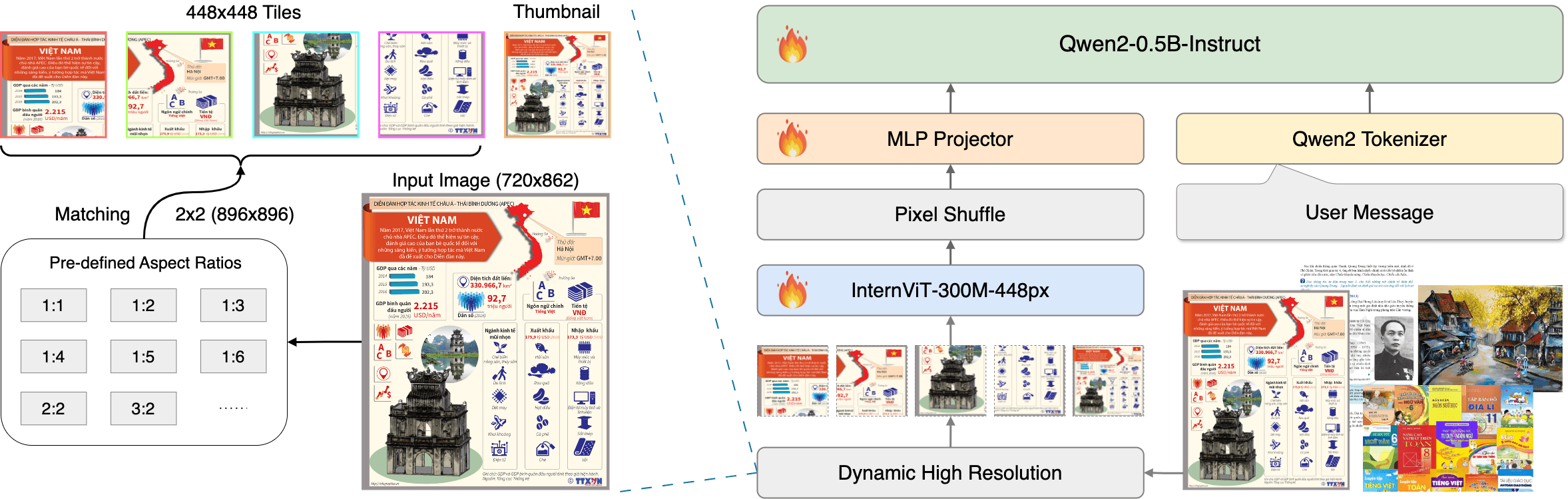}

\caption{Overall Architecture. Vintern-1B is built upon the ViT-MLP-LLM framework, following the structure of well-known MLLMs (\cite{chen2024internvl} \cite{liu2023llava} \cite{liu2023improvedllava} \cite{liu2024improvedllava1.6}). It inherits from InternVL 1.5 \cite{chen2024internvl}, integrating a pre-trained InternViT-300M-448px \cite{chen2024internvl} with Qwen2-0.5B-Instruct \cite{yang2024qwen2} via an MLP projector. The input image is processed by the Dynamic High Resolution module, which splits it into smaller 448x448 pixel images along with a thumbnail. These images are then passed through InternViT-300M-448px to extract visual features. A Pixel Shuffle step is also applied before feeding the data into the MLP projector to align it with the embeddings of the large language model Qwen2-0.5B-Instruct which takes the aligned visual tokens and the related question as inputs, and generates the corresponding answer.}


\label{im:internvl_arc}
\end{figure}

The Vintern-1B model is constructed using the InternVL-1.5 \ref{im:internvl_arc} architecture, which is similar to the 'ViT-MLP-LLM' configuration commonly found in many open-source MLLMs, as discussed in several studies. For more details, refer to Figure 1.

\begin{itemize}
    \item \textbf{Vision Encoder}: InternViT‑300M‑448px \cite{chen2024internvl} a distilled small vision foundation model, is used as the vision encoder, which transform the image to visual features.
    \item \textbf{MLP Projector}: A two-layer Multi-Layer Perceptron (MLP) projector is utilized to map the output representations from both the visual and language modalities into a unified space, facilitating the integration of visual and textual information.
    \item \textbf{Large Language Model}: We utilize the pre-trained Qwen2‑0.5B‑Instruct \cite{yang2024qwen2} serves as our language model, which generates text by leveraging the aligned representations provided by the MLP projector.
\end{itemize}

Our training processs follows the InternVL 1.5 \cite{chen2024internvl}, where images are segmented into $448\times448$ pixel tiles. We use pixel shuffle to reduce token count, enabling efficient high-resolution scaling, each $448\times448$ image is encoded as 256 visual tokens within the model.

\subsection{Vietnamese VQA Dataset}
\label{subset:datasets_desc}
With acknowledgment of the difference between Vietnamese images and IntervnVL's images. To enhance the Vintern-1B model's performance in the Vietnamese context, we extended several multimodal datasets. These datasets were designed to cover a wide range of tasks, from detailed image descriptions to OCR and document understanding, ensuring that the model is well-equipped to handle the nuances of Vietnamese visual and textual data.


\begin{table}[!b]
\caption{Datasets used in the fine-tuning stage.}
\label{table:dataset}
\centering
\begin{tabular}{l|l} 
\
\textbf{Task} & \textbf{Datasets} \\ 
\hline 
General QA & Vista \cite{ViVLMVista2024}, Viet-OpenViVQA-gemini-VQA, Viet-Localization-VQA \\
\hline
OCR & Viet-OCR-VQA, Viet-ViTextVQA-gemini-VQA, \\ & Viet-Vintext-gemini-VQA \\
\hline
Document & Viet-Doc-VQA, Viet-Doc-VQA-II, Viet-Geometry-VQA, \\
& Viet-ComputerScience-VQA, Viet-Sketches-VQA \\
\hline
Handwriting & Viet-Handwriting-VQA, Viet-Vintext-gemini-VQA \\
\hline
Extraction & Viet-Receipt-VQA, Viet-Menu-Gemini-VQA \\

\end{tabular}

\end{table}

\subsubsection{Vietnamese Image Extension}
We extended the Vietnamese multimodal datasets by collecting images from diverse sources, including web crawls across various topics relevant to Vietnamese culture and environment. These images include a wide array of content such as street scenes, traditional events, educational materials, and everyday objects, all of which reflect the unique characteristics of Vietnamese society. This extension is crucial for training the model to recognize and process visual information that is culturally specific to Vietnam. The
datasets used in fine-tuning phase are summarized in Table \ref{table:dataset}.

\subsubsection{Prompting with Gemini 1.5 Flash}
For each dataset, we utilized the Gemini 1.5 Flash model \cite{reid2024gemini} to generate concise and detailed descriptions, extract relevant information from the images, and create conversation-based query questions. This approach helps in enhancing the Vintern-1B model's ability to perform in various tasks within the Vietnamese context. By prompting Gemini 1.5 Flash, we ensured that the annotations were not only accurate but also aligned with the linguistic and cultural specifics of Vietnamese.

\subsubsection{General Question Answering (QA)}
To train the model in handling diverse QA scenarios, we included datasets such as:
\begin{itemize}
    \item \textbf{Vista \cite{ViVLMVista2024}:} A comprehensive dataset that covers a wide range of question types and contexts, designed to improve the model's understanding and generation of responses in Vietnamese.
    \item \textbf{Viet-OpenViVQA-gemini-VQA}: based on OpenViVQA \cite{nguyen2023openvivqa}: This dataset focuses on open-domain question answering, providing the model with exposure to various types of queries, from straightforward factual questions to more complex, context-dependent ones.
    \item \textbf{Viet-Localization-VQA}: includes quintessentially Vietnamese images such as scenic landscapes, historical sites, culinary specialties, festivals, cultural aspects from various regions, familiar rural scenes, and everyday life in urban areas, among others.
\end{itemize}

\begin{longtable}{|m{5cm}|m{5.5cm}|m{5.0cm}|}
    \caption{Samples in the Viet-OpenViVQA-gemini-VQA dataset and Viet-Localization-VQA dataset.}
    \label{tab:Viet-OpenViVQA-gemini-VQA-samples} \\
    \hline
    \textbf{Image} & \textbf{Description} & \textbf{Conversations} \\
    \hline
    \endfirsthead

    \hline
    \textbf{Image} & \textbf{Description} & \textbf{Conversations} \\
    \hline
    \endhead

    \hline
    \endfoot

    \hline
    \endlastfoot

    \hline

\includegraphics[width=5cm]{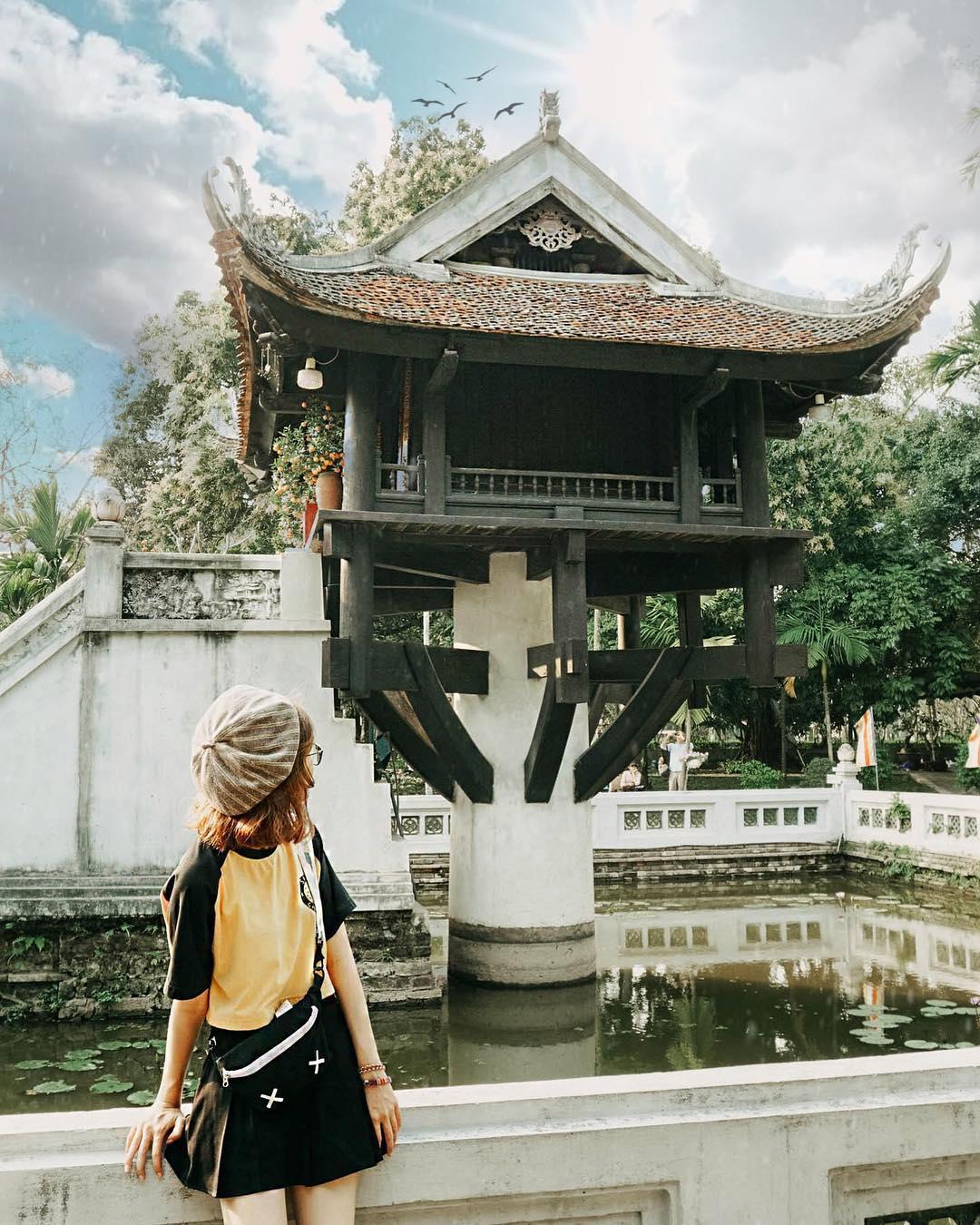} & \fontsize{8.0pt}{10pt}\selectfont{Bức ảnh chụp một cô gái trẻ đang đứng nhìn về phía Chùa Một Cột. Cô gái mặc áo màu vàng đen, chân váy đen, đội mũ nồi. Cô ấy đang đứng dựa vào lan can, phía sau là một hồ nước trong xanh, có rất nhiều hoa sen. Trên nền trời xanh có vài chú chim bay lượn. Chùa Một Cột được xây dựng trên một bệ đá hình trụ tròn, với kiến trúc độc đáo, mang đậm nét văn hóa truyền thống Việt Nam.

    \hfill \break
    \{The photo captures a young girl standing and looking towards the One Pillar Pagoda. She is wearing a yellow and black top, a black skirt, and a beret. She is leaning against a railing, with a clear blue lake full of lotuses behind her. In the blue sky, a few birds are flying. The One Pillar Pagoda is built on a round stone pillar, with unique architecture that deeply reflects traditional Vietnamese culture.\}
    
    } &
    \fontsize{8.0pt}{10pt}\selectfont{Q: Phía sau cô gái này là địa danh nào ?
    
    A: Phía sau cô gái này là Chùa Một Cột.
    
    Q: Cô gái đứng nhìn Chùa Một Cột mặc chiếc áo màu gì ?
    
    A: Cô ấy mặc chiếc áo màu vàng và đen
    
    Q: Cô gái đang quay đầu nhìn cái gì ?
    
    A: Cô ấy đang quay đầu nhìn Chùa Một Cột.
    
    \hfill \break
    \{Q: What landmark is behind the girl?
    
A: The One Pillar Pagoda is behind the girl.

Q: What color is the girl's top as she looks at the One Pillar Pagoda?

A: She is wearing a yellow and black top.

Q: What is the girl turning her head to look at?

A: She is turning her head to look at the One Pillar Pagoda.\}} \\
    \hline

\includegraphics[width=5cm]{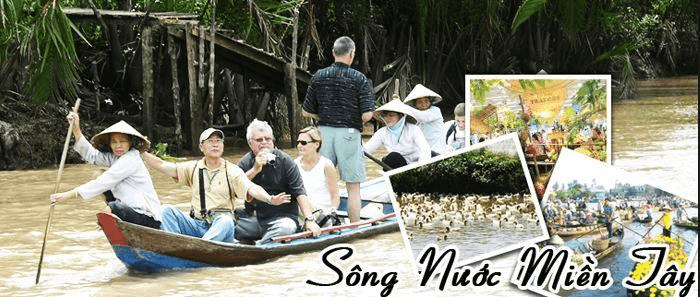} & 
    \fontsize{8.0pt}{10pt}\selectfont{Bức ảnh chụp cảnh sông nước miền Tây với một nhóm người đang ngồi trên một con thuyền. Bức ảnh được chia thành 3 phần chính. Phần trên là khung cảnh sông nước với một chiếc cầu gỗ bắc ngang qua sông. Phần giữa là hình ảnh những người đang ngồi trên thuyền, trong đó có một người phụ nữ đang chèo thuyền. Phần dưới là dòng chữ "Sông Nước Miền Tây" được viết theo kiểu chữ nghệ thuật. Hình ảnh được trang trí thêm 3 hình ảnh nhỏ ở góc phải, hình ảnh phía trên cùng là một chợ bán trái cây, hình ảnh bên dưới là một đàn vịt đang bơi trên sông và hình ảnh cuối cùng là một bến sông với những chiếc thuyền đậu dọc bờ sông. 

    \hfill \break
    \{A photograph of the river scene in the Mekong Delta with a group of people sitting on a boat. The picture is divided into three main parts. The top part shows the river landscape with a wooden bridge crossing the river. The middle part depicts people sitting on the boat, including a woman rowing. The bottom part features the words 'Mekong Delta River' written in an artistic font. The image is further decorated with three small pictures in the right corner: the topmost image shows a market selling fruits, the middle image shows a flock of ducks swimming in the river, and the final image shows a riverbank with boats moored along the shore.\}} & 
    \fontsize{8.0pt}{10pt}\selectfont{
    Q: Đây là khung cảnh ở đâu ?
    
    A: Đây là khung cảnh ở sông nước miền Tây.
    
    Q: Những người này đang ngồi ở đâu ?
    
    A: Những người này đang ngồi ở trên một con thuyền.
    
    Q: Những người phụ nữ đội nón lá đang làm gì ?
    
    A: Những người phụ nữ đội nón lá đang chèo thuyền.
    
    \hfill \break
    \{Q: Where is this scene?
    
    A: This is a scene of the Mekong Delta River.
        
    Q: Where are these people sitting?
    
    A: These people are sitting on a boat.
        
    Q: What are the women wearing conical hats doing?
    
    A: The women wearing conical hats are rowing the boat.\}} \\
    \hline

    \includegraphics[width=5cm]{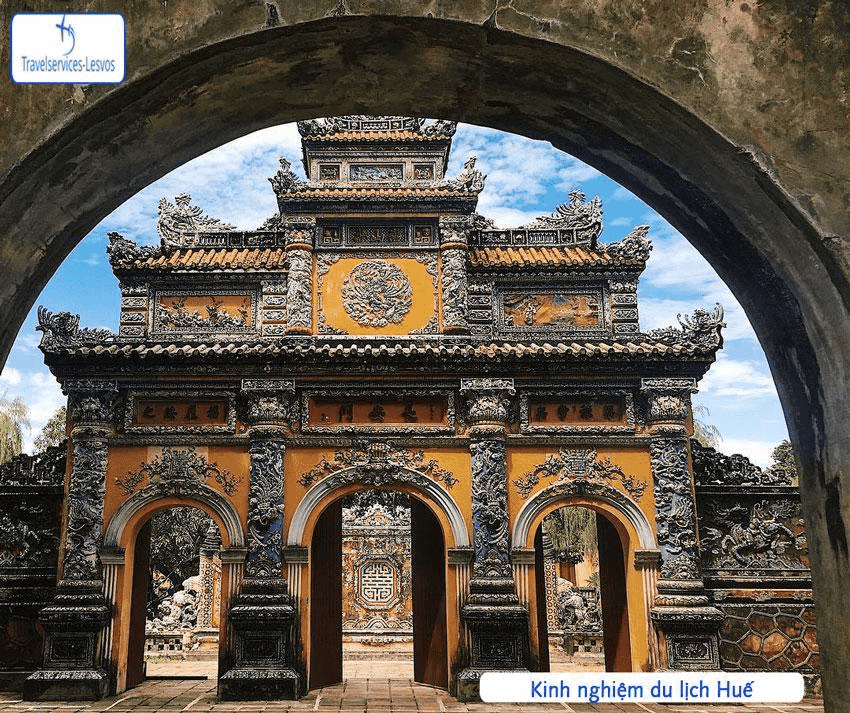} & 
    \fontsize{8.0pt}{10pt}\selectfont{Bức ảnh chụp cận cảnh một cổng được trang trí cầu kỳ, có màu vàng chủ đạo và điểm xuyết bằng những họa tiết màu xanh dương và đỏ. Cổng có ba lối vào được tạo thành bởi ba khung cửa hình vòm, được trang trí những đường nét hoa văn tinh xảo. Phía trước cổng có bốn cái cột trụ vững chắc, sơn màu vàng và được trang trí họa tiết rồng. Trên đỉnh cổng có nhiều mái cong vút, được trang trí bằng những họa tiết trang trí rực rỡ. Phía trên cổng, có chữ “Kinh nghiệm du lịch Huế” được in trên nền trắng. Bức ảnh được chụp từ phía sau một cổng vòm, tạo hiệu ứng như một khung ảnh, bao trọn khung cảnh của cổng và cảnh vật xung quanh. 

    \hfill \break
    \{The photo captures a close-up of an elaborately decorated gate, predominantly yellow with blue and red accents. The gate features three entrances formed by three arched doorways, adorned with intricate patterns. In front of the gate, there are four sturdy pillars, painted yellow and decorated with dragon motifs. At the top of the gate, several curved roofs are richly decorated with vibrant details. Above the gate, the words "Kinh nghiệm du lịch Huế" are printed on a white background. The photo is taken from behind an archway, creating a frame-like effect that encompasses the gate and the surrounding scenery.\}} & 
    \fontsize{8.0pt}{10pt}\selectfont{Q: Cái cổng thuộc địa phương nào ?
    
    A: Cái cổng thuộc Thành phố Huế
    
    Q: Cái cổng có bao nhiêu lối vào ?
    
    A: Cái cổng có ba lối vào
    
    Q: Phía trước cổng có bao nhiêu cái cột ?
    
    A: Phía trước cổng có bốn cái cột
    
    \hfill \break
    \{Q: Where is the gate located?
    
A: The gate is located in Hue City.

Q: How many entrances does the gate have?

A: The gate has three entrances.

Q: How many pillars are in front of the gate?

A: There are four pillars in front of the gate.\}
    } \\
    \hline
    \includegraphics[width=5cm]{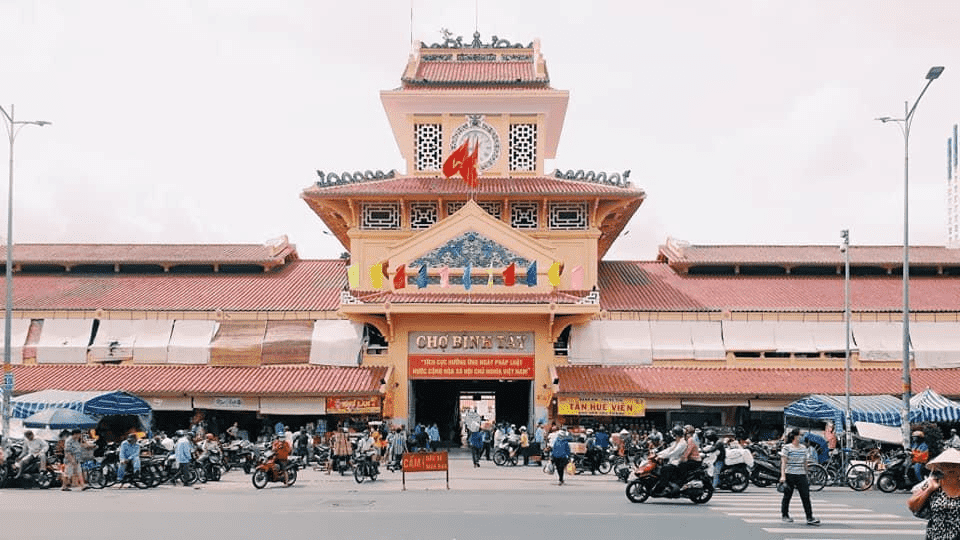} & 
    \fontsize{8.0pt}{10pt}\selectfont{Bức ảnh chụp toàn cảnh chợ Bình Tây từ phía trước. Chợ có kiến trúc đặc trưng với mái ngói đỏ và những họa tiết trang trí truyền thống. Phía trên cổng chính là tấm biển ghi chữ "CHỢ BÌNH TÂY" được in hoa. Phía dưới tấm biển là dòng chữ nhỏ "Tích cực hưởng ứng đề án phát huy giá trị di sản văn hóa của khu chợ Bình Tây" được in thường. Bên phải cổng chợ là một cửa tiệm có biển hiệu "TÂN HUÊ VIÊN" được in hoa. Trước cổng chợ là tấm biển cấm đậu xe mua bán. Ngoài ra, xung quanh chợ có rất nhiều người qua lại, xe máy, hàng quán và các gian hàng buôn bán.

    \hfill \break
    \{The photo captures a full view of Binh Tay Market from the front. The market features distinctive architecture with red tiled roofs and traditional decorative details. Above the main gate is a sign with the words "CHỢ BÌNH TÂY" in uppercase letters. Below the sign is a smaller line of text that reads "Tích cực hưởng ứng đề án phát huy giá trị di sản văn hóa của khu chợ Bình Tây" in lowercase letters. To the right of the market gate is a shop with a sign that reads "TÂN HUÊ VIÊN" in uppercase letters. In front of the market gate is a sign prohibiting parking for commercial activities. Additionally, the area around the market is bustling with people, motorcycles, shops, and various stalls.\}
    } & 
    \fontsize{8.0pt}{10pt}\selectfont{Q: Cửa tiệm bên phải chợ Bình Tây tên gì ?
    
    A: Cửa tiệm bên phải chợ Bình Tây tên là Tân Huê Viên.
    
    Q: Đây là ở đâu ?
    
    A: Đây là chợ Bình Tây.
    
    Q: Tấm biển trước chợ Bình Tây ghi gì ?
    
    A: Tấm biển trước chợ Bình Tây ghi là Cấm đậu xe mua bán.

    \hfill \break
    \{Q: What is the name of the shop to the right of Binh Tay Market?
A: The shop to the right of Binh Tay Market is called "Tân Huê Viên".

Q: Where is this?
A: This is Binh Tay Market.

Q: What does the sign in front of Binh Tay Market say?
A: The sign in front of Binh Tay Market says "Cấm đậu xe mua bán"\}

    } \\
    \hline
    \includegraphics[width=5cm]{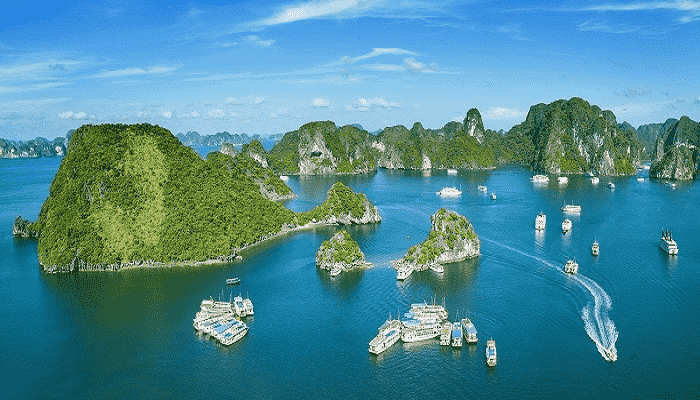} & 
    \fontsize{8.0pt}{10pt}\selectfont{Hình ảnh cho thấy một khung cảnh thiên nhiên tuyệt đẹp với biển xanh, trời trong, nắng đẹp, mây trắng bay bổng. Nổi bật trên mặt biển là vô số hòn đảo lớn nhỏ với nhiều hình thù khác nhau, được bao phủ bởi một màu xanh tươi mát của cây cối. Các đảo có hình dáng nhô lên từ mặt nước, tạo thành những vách đá dựng đứng, một số hòn đảo lại có hình dáng tròn trịa, thấp thoáng những bãi cát trắng mịn.  Đặc biệt, ở trung tâm của bức ảnh, một hòn đảo lớn có diện tích khá rộng, được bao phủ bởi rừng cây xanh um tùm. Trên mặt biển, có nhiều thuyền bè, tàu thuyền di chuyển, tạo nên một khung cảnh nhộn nhịp, sôi động. Nước biển có màu xanh ngọc bích, phản chiếu ánh nắng mặt trời tạo nên một vẻ đẹp lung linh, huyền ảo. Xa xa, có thể nhìn thấy những dãy núi trùng điệp, ẩn hiện trong màn sương mờ ảo, tạo nên một khung cảnh hùng vĩ, thơ mộng.

    \hfill \break
    \{The image shows a beautiful natural landscape with blue sea, clear sky, beautiful sunshine, and floating white clouds. Standing out on the sea are countless large and small islands with many different shapes, covered by a fresh green color of trees. The islands have shapes that protrude from the water, forming steep cliffs, some islands have round shapes, with white sand beaches looming. In particular, in the center of the photo, a large island with a fairly large area is covered by lush green forests. On the sea, there are many boats and ships moving, creating a bustling, vibrant scene. The sea water is emerald green, reflecting the sunlight, creating a shimmering, magical beauty. In the distance, you can see the overlapping mountain ranges, hidden in the misty mist, creating a majestic, poetic scene.\}
    } & 
    \fontsize{8.0pt}{10pt}\selectfont{Q: Hình ảnh cho thấy có bao nhiêu hòn đảo và đặc điểm nổi bật của chúng là gì?
    
A: Hình ảnh cho thấy có rất nhiều hòn đảo lớn nhỏ với nhiều hình thù khác nhau. Chúng được bao phủ bởi một màu xanh tươi mát của cây cối. Các đảo có hình dáng nhô lên từ mặt nước, tạo thành những vách đá dựng đứng, một số hòn đảo lại có hình dáng tròn trịa, thấp thoáng những bãi cát trắng mịn. Một hòn đảo lớn có diện tích khá rộng, được bao phủ bởi rừng cây xanh um tùm.

    \hfill \break
    \{Q: How many islands are there in the picture and what are their outstanding features?

A: The picture shows many large and small islands with many different shapes. They are covered with a fresh green color of trees. The islands have shapes that protrude from the water surface, forming steep cliffs, some islands have round shapes, with white sand beaches looming. A large island has a fairly large area, covered with lush green forests.

\}

    } \\
    \hline
    
\end{longtable}

\vfil \break       
\subsubsection{OCR and Text Recognition}
For OCR-related tasks, where the goal is to recognize and interpret text from images, we used:
\begin{itemize}
    \item \textbf{Viet-ViTextVQA-gemini-VQA} based on ViTextVQA \cite{van2024vitextvqa}: A dataset specifically designed to improve the model's ability to recognize Vietnamese text in various image contexts, such as signs, documents, and digital screens.
    \item \textbf{Viet-Vintext-gemini-VQA} based on Vintext \cite{nguyen2021dictionaryvintext}: This dataset contains a rich collection of Vietnamese scene text images, aimed at enhancing text recognition capabilities in real-world.
    \item \textbf{Viet-OCR-VQA}: We introduced this new large-scale dataset, detailed in our Hugging Face repository. It includes over 137,000 images containing Vietnamese text, accompanied by more than 822,679 description and query-answer pairs. This dataset is particularly valuable for tasks involving scene text recognition and understanding.
\end{itemize}

\begin{longtable}{|m{5cm}|m{6.0cm}|m{4.5cm}|}
    \caption{Samples in the Viet-OCR-VQA dataset.}
    \label{tab:Viet-OCR-VQA-samples} \\
    
    \hline
    \textbf{Image} & \textbf{Description} & \textbf{Conversations} \\
    \hline
    \endfirsthead

    \hline
    \textbf{Image} & \textbf{Description} & \textbf{Conversations} \\
    \hline
    \endhead

    \hline
    \endfoot

    \hline
    \endlastfoot
    \includegraphics[width=5cm]{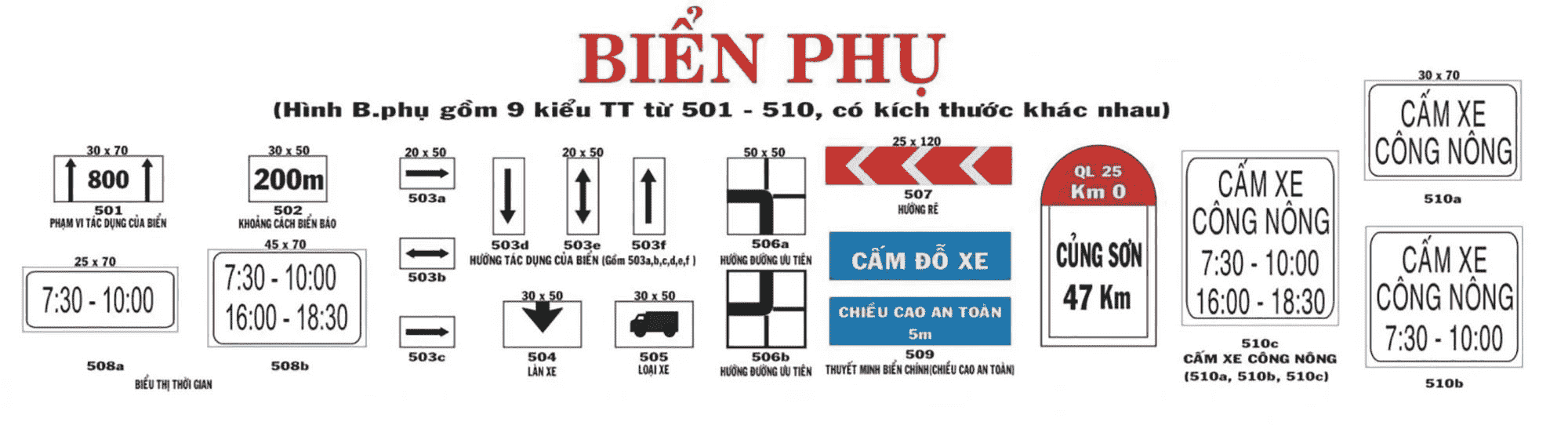} & 
    \fontsize{8.0pt}{10pt}\selectfont{Bức ảnh là sơ đồ minh họa về các loại biển báo giao thông phụ trong hệ thống biển báo giao thông Việt Nam, với tiêu đề "BIỂN PHỤ" được in chữ in hoa cỡ lớn và in đậm. Dưới tiêu đề là dòng chữ giải thích: "(Hình B.phụ gồm 9 kiểu TT từ 501 - 510, có kích thước khác nhau)". Phần chính của ảnh là 9 biển báo được sắp xếp theo hàng ngang với mỗi biển báo được đánh số thứ tự từ 501 đến 510, và được ghi chú thêm mã phụ a, b, c, d, e, f. Bên cạnh mỗi biển báo là kích thước của biển báo đó, ví dụ 30 x 70. Bên dưới mỗi biển báo là nội dung giải thích ngắn gọn về ý nghĩa của biển báo đó, ví dụ "PHẠM VI TÁC DỤNG CỦA BIỂN", "KHOẢNG CÁCH BIỂN BÁO", "HƯỚNG TÁC DỤNG CỦA BIỂN (Gồm 503a,b,c,d,e,f)", "HƯỚNG ĐƯỜNG ƯU TIÊN", "LOẠI XE", "THUYẾT MINH BIỂN CHÍNH (CHIỀU CAO AN TOÀN)". Ngoài ra, có hai biển báo nằm ở góc trên bên phải của ảnh, với nội dung "CẤM XE CÔNG NÔNG" và ghi chú thời gian cấm, ví dụ "7:30 - 10:00, 16:00 - 18:30".

    \hfill \break
    \{The image is a diagram illustrating the types of auxiliary traffic signs in the Vietnamese traffic sign system, with the title 'BIỂN PHỤ' printed in large, bold uppercase letters. Below the title is an explanatory line: "(Hình B.phụ gồm 9 kiểu TT từ 501 - 510, có kích thước khác nhau)". The main part of the image consists of 9 signs arranged horizontally, each numbered from 501 to 510, and annotated with additional subcodes a, b, c, d, e, f. Next to each sign is the size of that sign, for example, 30 x 70. Below each sign is a brief explanation of the meaning of that sign, for example, "PHẠM VI TÁC DỤNG CỦA BIỂN","KHOẢNG CÁCH BIỂN BÁO", "HƯỚNG TÁC DỤNG CỦA BIỂN (Gồm 503a,b,c,d,e,f)", "HƯỚNG ĐƯỜNG ƯU TIÊN", "LOẠI XE", "THUYẾT MINH BIỂN CHÍNH (CHIỀU CAO AN TOÀN)". Additionally, there are two signs in the top right corner of the image, with the content "CẤM XE CÔNG NÔNG" and a note on the prohibited time, for example, "7:30 - 10:00, 16:00 - 18:30".\}
    }
    & 
    \fontsize{8.0pt}{10pt}\selectfont{Q: Tiêu đề của sơ đồ minh họa về các biển báo phụ trong ảnh là gì
    
    A:Tiêu đề của sơ đồ là \"BIỂN PHỤ\".
    
    Q:Số lượng kiểu biển báo phụ được minh họa trong ảnh là bao nhiêu
    A: Có 9 kiểu biển báo phụ được minh họa trong ảnh.
    
    Q:Khoảng thời gian cấm xe công nông được ghi chú trên biển báo 510b là gì?
    
    A: Khoảng thời gian cấm xe công nông trên biển báo 510b là 7:30 - 10:00.
    
    Q:Biển báo số 503d có ý nghĩa gì
    
    A: Biển báo 503d có ý nghĩa là "HƯỚNG TÁC DỤNG CỦA BIỂN".
    
    Q:Kích thước của biển báo số 509 là bao nhiêu?
    
    A: Kích thước của biển báo số 509 là 25 x 120.

    \hfill \break
    \{Q: What is the title of the diagram illustrating the auxiliary signs in the image?

A: The title of the diagram is "BIỂN PHỤ".

Q: How many types of auxiliary signs are illustrated in the image?

A: There are 9 types of auxiliary signs illustrated in the image.

Q: What is the restricted time for tractors noted on sign 510b?

A: The restricted time for tractors on sign 510b is 7:30 - 10:00.

Q: What is the meaning of sign 503d?

A: Sign 503d means "HƯỚNG TÁC DỤNG CỦA BIỂN".

Q: What is the size of sign 509?

A: The size of sign 509 is 25 x 120.\}
    } \\
    \hline
    \includegraphics[width=5cm]{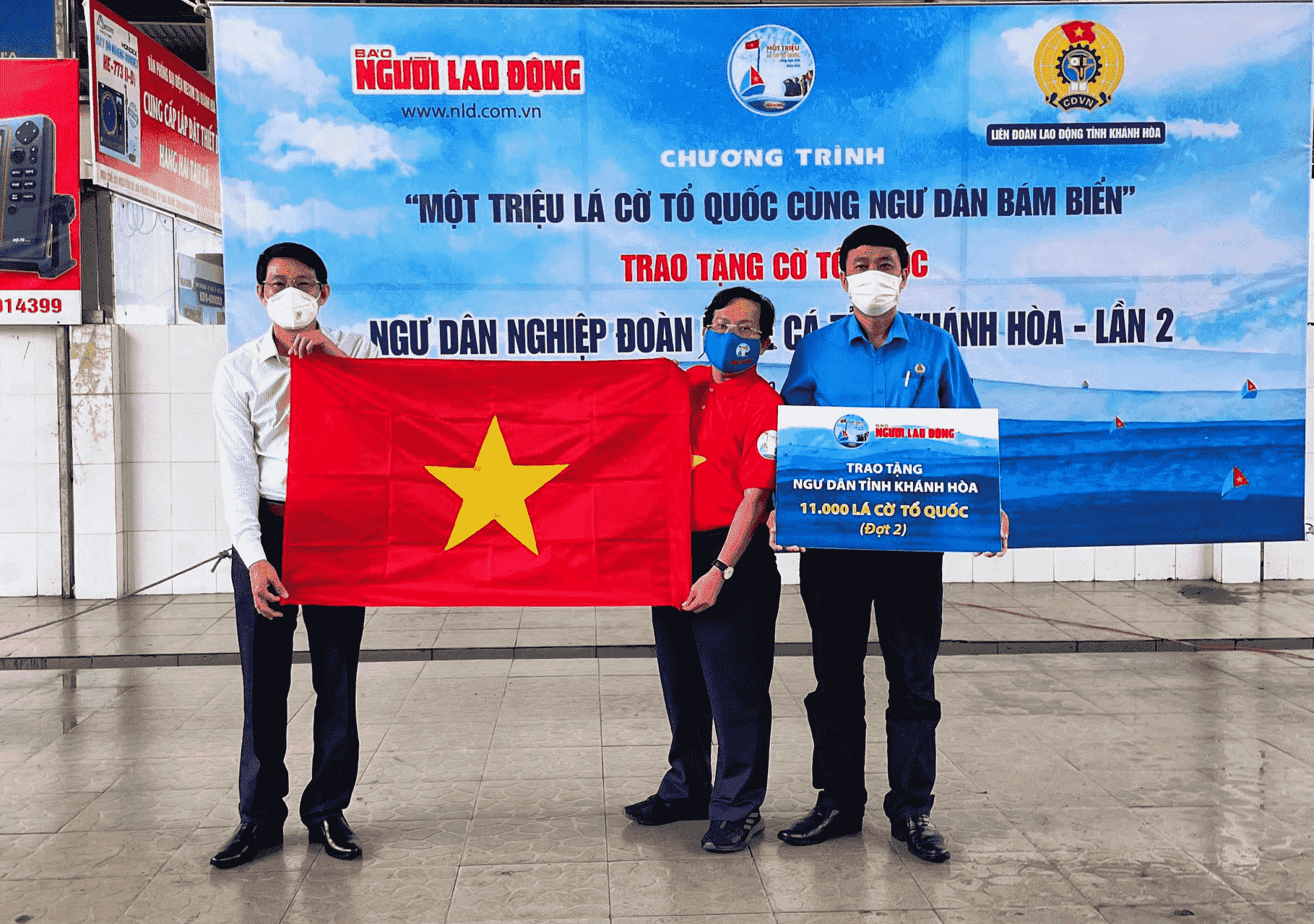} & 
    \fontsize{8.0pt}{10pt}\selectfont{Bức ảnh chụp một nhóm người đứng trước một tấm biển quảng cáo lớn có nền màu xanh dương. Trên biển có dòng chữ "Báo Người Lao Động" với logo và website www.nld.com.vn. Dưới đó là tiêu đề chương trình ''Một triệu là cờ tổ quốc cùng ngư dân bám biển" và ''Trao tặng cờ tổ quốc." Phía dưới là dòng chữ "Ngư dân nghiệp đoàn cá tỉnh Khánh Hoà - Lần 2''. Người đứng giữa cầm một tấm bảng màu trắng ghi ''Trao tặng ngư dân tỉnh Khánh Hòa 11.000 là cờ tổ quốc (Đợt 2)'' Người bên trái cầm cờ Việt Nam, còn người bên phải mặc áo blue có in chữ ''Báo Người Lao Động'' ở ngực. Ở góc trên bên phải của tấm biển có dòng chữ '' Liên đoàn Lao động tỉnh Khánh Hò'' và logo Liên đoàn Lao động.

    \hfill \break
    \{The photo captures a group of people standing in front of a large blue advertisement board. On the board, there is the text "Báo Người Lao Động" along with the logo and website www.nld.com.vn. Below that is the title of the program "Một triệu lá cờ tổ quốc cùng ngư dân bám biển" and "Trao tặng cờ tổ quốc." Beneath this is the text "Ngư dân nghiệp đoàn cá tỉnh Khánh Hòa - Lần 2." The person in the center is holding a white sign that reads "Trao tặng ngư dân tỉnh Khánh Hòa 11.000 lá cờ tổ quốc (Đợt 2)." The person on the left is holding a Vietnamese flag, while the person on the right is wearing a blue shirt with the text "Báo Người Lao Động" printed on the chest. In the top right corner of the board is the text "Liên đoàn Lao động tỉnh Khánh Hòa" and the logo of the Labor Federation.\}
    } & 
    \fontsize{8.0pt}{10pt}\selectfont{Q: Chương trình trong ảnh có tên gọi là gì?
    
    A: Chương trình trong ảnh có tên gọi là ''Một triệu là cờ tổ quốc cùng ngư dân bám biển''
    
    Q: Chương trình được tổ chức ở đâu?
    
    A: Chương trình được tổ chức tại tỉnh Khánh Hòa.
    
    Q: Chương trình trao tặng gì cho ngư dân?
    
    A: Chương trình trao tặng cờ tổ quốc cho ngư dân.
    
    Q: Số lượng cờ tổ quốc được trao tặng là bao nhiêu?
    
    A: Số lượng cờ tổ quốc được trao tặng là 11.000 lá cờ.
    
    Q: Chương trình được tổ chức ở lần thứ mấy?
    
    A: Chương trình được tổ chức ở lần thứ 2.
    
    \hfill \break
    \{Q: What is the name of the program in the photo?
    
A: The program in the photo is called "Một triệu lá cờ tổ quốc cùng ngư dân bám biển."

Q: Where was the program held?

A: The program was held in Khánh Hòa province.

Q: What does the program give to the fishermen?

A: The program gives national flags to the fishermen.

Q: How many national flags were given?

A: A total of 11,000 national flags were given.

Q: How many times has the program been held?

A: The program has been held for the second time.\}}\\
    \hline

\end{longtable}

\vfil \break

\subsubsection{Document Understanding}
To tackle document understanding tasks, which require the model to comprehend and extract information from structured documents, we developed:
\begin{itemize}
    \item \textbf{Viet-Doc-VQA \& Viet-Doc-VQA-II}: These datasets were created from over 116,000 pages of Vietnamese textbooks, covering subjects from grades 1 to 12, including Mathematics, Literature, English, Physics, Chemistry, Biology, History, Geography, Civic Education, Computer Science, Technology, Music, Arts, and Physical Education. These datasets not only facilitate real-world document analysis but also help the model learn and understand educational content relevant to Vietnam.
    \item \textbf{Viet-Geometry-VQA \& Viet-ComputerScience-VQA}: These datasets focus on scientific image understanding, particularly in math and logic problems, enabling the model to solve and interpret visual problems in these domains.
    \item \textbf{Viet-Sketches-VQA}: Designed to improve the model's ability to understand symbolic representations and sketches, this dataset aids in tasks where visual abstraction and symbol interpretation are required.
\end{itemize}

\begin{longtable}{|m{5cm}|m{5.5cm}|m{5cm}|}
    \caption{Samples in the Vietnamese document understanding datasets.}
    \label{tab:Viet-Document-Understading-VQA-samples} \\
    
    \hline
    \textbf{Image} & \textbf{Description} & \textbf{Conversations} \\
    \hline
    \endfirsthead

    \hline
    \textbf{Image} & \textbf{Description} & \textbf{Conversations} \\
    \hline
    \endhead

    \hline
    \endfoot

    \hline
    \endlastfoot

    \includegraphics[width=5cm]{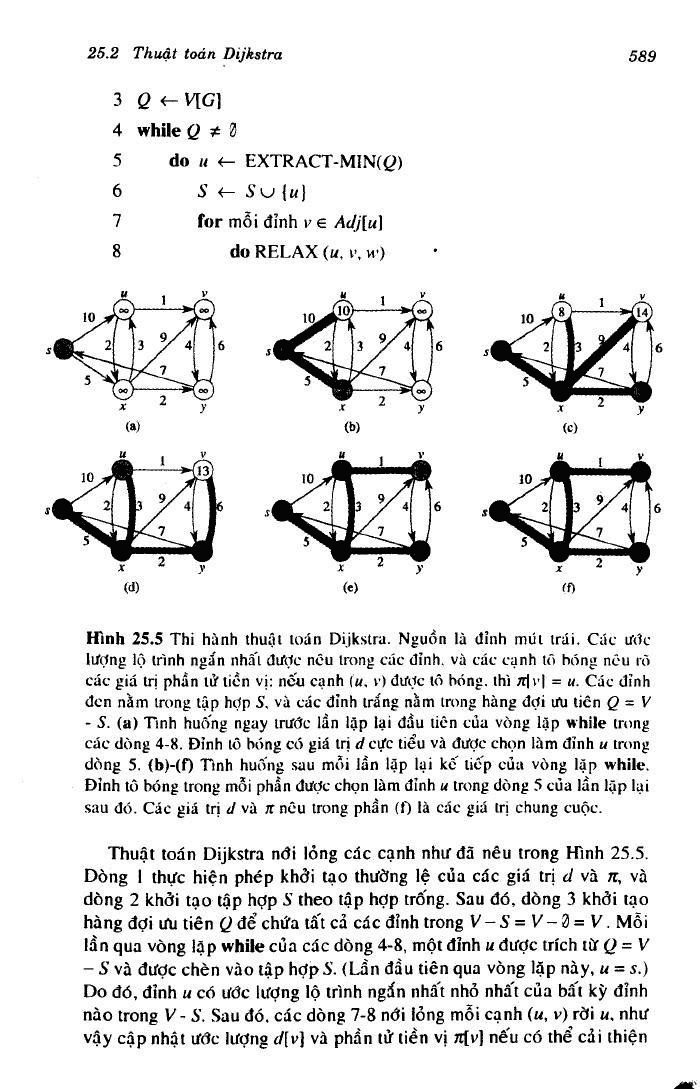} & 
    \fontsize{8pt}{10pt}\selectfont{Bức ảnh mô tả thuật toán Dijkstra tìm đường đi ngắn nhất trong đồ thị. Phần trên của ảnh là phần code thuật toán, với các bước thực hiện:
- Khởi tạo tập hợp đỉnh chưa thăm S và hàng đợi ưu tiên Q, Q bao gồm tất cả các đỉnh trong đồ thị.
- Lặp cho đến khi Q rỗng: - Lấy đỉnh u có trọng số d[u] nhỏ nhất khỏi Q - Thêm đỉnh u vào tập hợp S.
- Cập nhật trọng số d[v] của các đỉnh v kề với u nếu d[u] + trọng số cạnh (u,v) nhỏ hơn d[v].
Phần dưới của ảnh là hình minh họa cho từng bước thuật toán. Hình minh họa cho thấy đồ thị với các cạnh có trọng số,
các đỉnh được tô màu đen hoặc trắng. Đỉnh đen là những đỉnh đã được thêm vào tập S. - Hình (a) thể hiện đồ thị ban đầu.
- Các hình (b) đến (f) thể hiện quá trình thuật toán Dijkstra hoạt động. - Mỗi hình minh họa cho thấy các đỉnh được cập nhật, đỉnh có màu đen là đỉnh đã được thêm vào tập hợp S.
- Trọng số d[v] được hiển thị bên cạnh mỗi đỉnh. Phần giải thích ở dưới cùng giải thích chi tiết cách thuật toán hoạt động.

\hfill \break
\{Iterate until Q is empty: - Extract the vertex u with the smallest weight d[u] from Q - Add vertex u to the set S.
Update the weight d[v] of the vertices v adjacent to u if d[u] + the edge weight (u,v) is less than d[v].
The lower part of the image is an illustration of each step of the algorithm. The illustration shows a graph with weighted edges, where vertices are colored black or white. Black vertices are those that have been added to the set S.
Figure (a) shows the initial graph.
Figures (b) through (f) depict the process of Dijkstra's algorithm in action. - Each figure illustrates the vertices being updated, with black vertices representing those that have been added to the set S.
The weight d[v] is displayed next to each vertex. The explanation at the bottom provides detailed information on how the algorithm works.\}
} & 
\fontsize{8.0pt}{10pt}\selectfont{Q: Thuật toán Dijkstra giải quyết bài toán gì?

A: Thuật toán Dijkstra giải quyết bài toán tìm đường đi ngắn nhất từ một đỉnh nguồn đến các đỉnh khác trong đồ thị có trọng số. Thuật toán sử dụng phương pháp tham lam để tìm kiếm các cạnh có trọng số nhỏ nhất để tạo thành đường đi ngắn nhất.

Q: Trong hình minh họa, đỉnh nào là đỉnh nguồn?

A: Trong hình minh họa, đỉnh nguồn là đỉnh s, được tô màu đen ở hình (a).

Q: Tại sao các đỉnh được tô màu đen trong các hình (b) đến (f)?

A: Các đỉnh được tô màu đen trong các hình (b) đến (f) là các đỉnh đã được thêm vào tập hợp S, tức là các đỉnh đã được thuật toán Dijkstra tìm đường đi ngắn nhất từ đỉnh nguồn đến. Các đỉnh này đã được xử lý và trọng số d[v] của chúng đã được cập nhật.

Q: Thuật toán Dijkstra hoạt động dựa trên nguyên tắc nào?

A: Thuật toán Dijkstra hoạt động dựa trên nguyên tắc tham lam. Nó sẽ luôn chọn cạnh có trọng số nhỏ nhất để di chuyển từ đỉnh hiện tại đến đỉnh kế tiếp. Do đó, thuật toán này sẽ đảm bảo rằng đường đi ngắn nhất được tìm thấy.

\hfill \break
\{Q: What problem does Dijkstra's algorithm solve?

A: Dijkstra's algorithm solves the problem of finding the shortest path from a source vertex to other vertices in a weighted graph. The algorithm uses a greedy approach to find the edges with the smallest weights to form the shortest path.

Q: In the illustration, which vertex is the source vertex?

A: In the illustration, the source vertex is vertex s, colored black in figure (a).

Q: Why are the vertices colored black in figures (b) to (f)?

A: The vertices colored black in figures (b) to (f) are the vertices that have been added to the set S, meaning they are the vertices for which Dijkstra's algorithm has found the shortest path from the source vertex. These vertices have been processed, and their weight d[v] has been updated.

Q: On what principle does Dijkstra's algorithm operate?

A: Dijkstra's algorithm operates on the principle of greediness. It always selects the edge with the smallest weight to move from the current vertex to the next vertex. Therefore, this algorithm ensures that the shortest path is found.\}
} \\

    \hline
    \includegraphics[width=5cm]{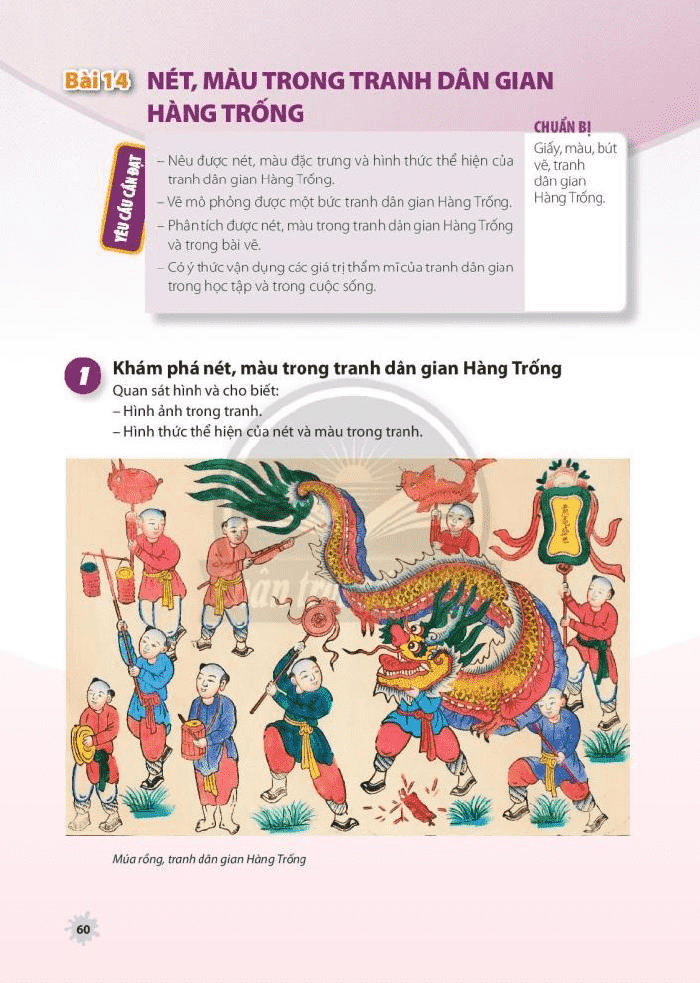} & 
    \fontsize{8.0pt}{10pt}\selectfont{Bức ảnh được chia thành hai phần: phần trên và phần dưới. Phần trên là tiêu đề bài học với nội dung 'Nét, màu trong tranh dân gian Hàng Trống'. Phần trên được chia thành hai cột, cột bên trái là tiêu đề 'Bài 14' và cột bên phải bao gồm phần nội dung giải thích về nét, màu sắc và hình thức thể hiện trong tranh dân gian Hàng Trống. Dòng đầu tiên của phần nội dung giải thích 'Nêu được nét, màu đặc trưng và hình thức thể hiện của tranh dân gian Hàng Trống'. Dòng tiếp theo là 'Vẽ mô phỏng được một bức tranh dân gian Hàng Trống'. Dòng thứ ba là 'Phân tích được nét, màu trong tranh dân gian Hàng Trống và trong bài vẽ'. Dòng cuối cùng là 'Có ý thức vận dụng các giá trị thẩm mĩ của tranh dân gian trong học tập và trong cuộc sống'. Bên cạnh cột nội dung là một khung hình chữ nhật màu hồng với tiêu đề 'YÊU CẦU CẦN ĐẠT', bên cạnh cột nội dung là một khung hình chữ nhật màu vàng với tiêu đề 'CHUẨN BỊ', nội dung bên trong khung là 'Giấy, màu, bút vẽ, tranh dân gian Hàng Trống'. Phần dưới là nội dung 'Khám phá nét, màu trong tranh dân gian Hàng Trống', dòng tiếp theo là 'Quan sát hình và cho biết', bên cạnh là số '1' được đặt trong một khung hình tròn màu xanh lá cây. Bên dưới có ba dòng chữ 'Hình ảnh trong tranh', 'Hình thức thể hiện của nét và màu trong tranh', 'Mùa rồng, tranh dân gian Hàng Trống'.

    \hfill \break
    \{The image is divided into two parts: the upper part and the lower part. The upper part is the lesson title with the content 'Nét, màu trong tranh dân gian Hàng Trống'. The upper part is divided into two columns; the left column has the title 'Bài 14', and the right column includes the content explaining the lines, colors, and forms of expression in Hàng Trống folk paintings. The first line of the explanatory content states, 'Nêu được nét, màu đặc trưng và hình thức thể hiện của tranh dân gian Hàng Trống'. The next line is 'Vẽ mô phỏng được một bức tranh dân gian Hàng Trống'. The third line is 'Phân tích được nét, màu trong tranh dân gian Hàng Trống và trong bài vẽ'. The last line is 'Có ý thức vận dụng các giá trị thẩm mĩ của tranh dân gian trong học tập và trong cuộc sống'. Next to the content column is a pink rectangular box with the title 'YÊU CẦU CẦN ĐẠT', and beside it is a yellow rectangular box with the title 'CHUẨN BỊ', with the content inside the box being 'Giấy, màu, bút vẽ, tranh dân gian Hàng Trống'. The lower part is the content 'Khám phá nét, màu trong tranh dân gian Hàng Trống', the next line is 'Quan sát hình và cho biết', with the number '1' placed in a green circular frame beside it. Below that, there are three lines of text: 'Hình ảnh trong tranh', 'Hình thức thể hiện của nét và màu trong tranh', 'Mùa rồng, tranh dân gian Hàng Trống'.\}
    } & 
    \fontsize{8.0pt}{10pt}\selectfont{Q: Trong tranh dân gian Hàng Trống, hình ảnh nào được thể hiện? 
    
    A: Tranh dân gian Hàng Trống thể hiện hình ảnh múa rồng. Hình ảnh rồng được thể hiện bằng nét vẽ đơn giản, màu sắc rực rỡ. Rồng được miêu tả với nhiều màu sắc khác nhau như vàng, đỏ, xanh lá cây. Ngoài ra, tranh còn có hình ảnh các nhân vật đang múa rồng. Các nhân vật được thể hiện bằng nét vẽ đơn giản, màu sắc rực rỡ. Nhân vật được thể hiện với nhiều tư thế khác nhau, tạo nên sự sinh động cho bức tranh. 
    
    Q: Nét, màu sắc và hình thức thể hiện của tranh dân gian Hàng Trống như thế nào? 
    
    A: Nét, màu sắc và hình thức thể hiện của tranh dân gian Hàng Trống được thể hiện bằng nét vẽ đơn giản, màu sắc rực rỡ. Màu sắc chủ yếu là màu đỏ, vàng, xanh lá cây. Hình thức thể hiện của tranh dân gian Hàng Trống là hình thức thể hiện với nét vẽ đơn giản, không cầu kỳ, và màu sắc rực rỡ, tạo nên sự vui tươi, sinh động cho bức tranh. 

    \hfill \break
    \{Q: In Hàng Trống folk paintings, which image is depicted?

A: The Hàng Trống folk painting depicts the image of a dragon dance. The dragon image is represented with simple lines and vibrant colors. The dragon is described with various colors such as yellow, red, and green. Additionally, the painting includes images of characters performing the dragon dance. The characters are depicted with simple lines and vibrant colors, portrayed in different poses, adding liveliness to the painting.

Q: How are the lines, colors, and forms of expression in Hàng Trống folk paintings?

A: The lines, colors, and forms of expression in Hàng Trống folk paintings are represented by simple lines and vibrant colors. The main colors used are red, yellow, and green. The form of expression in Hàng Trống folk paintings is characterized by simple lines, not overly intricate, and vibrant colors, creating a joyful and lively atmosphere in the painting.
\}
    } \\
    \hline

    \includegraphics[width=5cm]{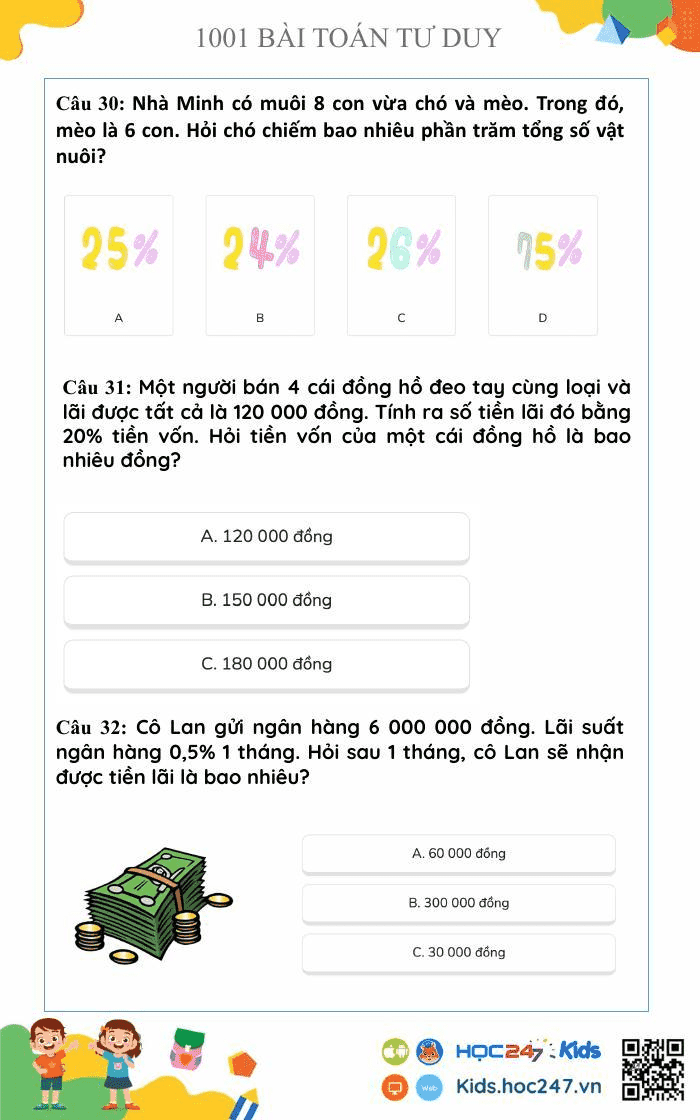} & 
    \fontsize{8.0pt}{10pt}\selectfont{Bức ảnh hiển thị 3 câu hỏi trắc nghiệm với 4 đáp án được đánh dấu A, B, C, D cho mỗi câu. Câu 30 hỏi về tỷ lệ phần trăm chó trong tổng số vật nuôi, minh họa bằng 4 hình tròn, mỗi hình tròn chứa một đáp án phần trăm (25\%, 24\%, 26\%, 75\%). Câu 31 hỏi về tiền vốn của một chiếc đồng hồ, minh họa bằng 3 ô chữ nhật, mỗi ô chứa một đáp án tiền (120 000 đồng, 150 000 đồng, 180 000 đồng). Câu 32 hỏi về tiền lãi nhận được sau một tháng, minh họa bằng hình ảnh một chồng tiền và 3 ô chữ nhật, mỗi ô chứa một đáp án tiền (60 000 đồng, 300 000 đồng, 30 000 đồng).

    \hfill \break
    \{The picture displays 3 multiple-choice questions with 4 answer options labeled A, B, C, D for each question. Question 30 asks about the percentage of dogs in the total number of pets, illustrated by 4 circles, each containing a percentage answer (25\%, 24\%, 26\%, 75\%). Question 31 asks about the capital for a watch, illustrated by 3 rectangles, each containing a monetary answer (120,000 VND, 150,000 VND, 180,000 VND). Question 32 asks about the interest earned after one month, illustrated by an image of a stack of money and 3 rectangles, each containing a monetary answer (60,000 VND, 300,000 VND, 30,000 VND).\}
    } & 
    \fontsize{8.0pt}{10pt}\selectfont{Q: Trong câu 30, tổng số vật nuôi là bao nhiêu? 
    
    A: Câu 30 cho biết tổng số vật nuôi là 8 con, bao gồm cả chó và mèo. 
    
    Q: Theo hình minh họa câu 30, đáp án nào thể hiện tỷ lệ chó nhiều nhất?
    
    A: Theo hình minh họa, đáp án D (75\%) thể hiện tỷ lệ chó nhiều nhất. 
    
    Q: Câu 31 hỏi về tiền vốn của một chiếc đồng hồ. Vậy, tổng số tiền lãi là bao nhiêu? A: Câu 31 cho biết tổng số tiền lãi là 120 000 đồng. 
    
    Q: Trong câu 32, số tiền gửi ngân hàng là bao nhiêu? 
    
    A: Câu 32 cho biết số tiền gửi ngân hàng là 6 000 000 đồng. 
    
    Q: Theo hình minh họa câu 32, đáp án nào thể hiện số tiền lãi lớn nhất? 
    
    A: Theo hình minh họa, đáp án B (300 000 đồng) thể hiện số tiền lãi lớn nhất.

    \hfill \break
    \{Q: In question 30, what is the total number of pets?

A: Question 30 states that the total number of pets is 8, including both dogs and cats.

Q: According to the illustration in question 30, which answer shows the highest percentage of dogs?

A: According to the illustration, answer D (75

Q: Question 31 asks about the capital for a watch. So, what is the total interest?

A: Question 31 states that the total interest is 120,000 VND.

Q: In question 32, how much money is deposited in the bank?

A: Question 32 states that the amount of money deposited in the bank is 6,000,000 VND.

Q: According to the illustration in question 32, which answer shows the highest interest?

A: According to the illustration, answer B (300,000 VND) shows the highest interest.\}
    
    } \\
    \hline
    \includegraphics[width=5cm]{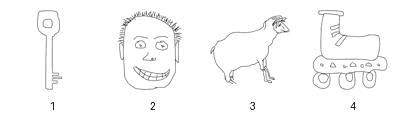} & 
    \fontsize{8pt}{10pt}\selectfont{Bức ảnh được chia thành bốn phần bằng nhau, mỗi phần chứa một hình vẽ được đánh số thứ tự từ 1 đến 4. Hình vẽ số 1 là một chiếc chìa khóa với phần đầu tròn, phần thân thẳng và có rãnh ở cuối. Hình vẽ số 2 là một khuôn mặt người, với mái tóc rối, mắt to, mũi ngắn và miệng cười rộng. Hình vẽ số 3 là một con dê, với phần đầu nghiêng, sừng cong, thân hình tròn và chân gầy. Hình vẽ số 4 là một đôi giày trượt patin, với phần thân cao, dây buộc ngang và bánh xe tròn. 

    \hfill \break
    \{The picture is divided into four equal parts, each containing a drawing numbered from 1 to 4. Drawing number 1 is a key with a round head, a straight shaft, and a groove at the end. Drawing number 2 is a human face with messy hair, large eyes, a short nose, and a wide smiling mouth. Drawing number 3 is a goat with a tilted head, curved horns, a round body, and skinny legs. Drawing number 4 is a pair of roller skates with a tall body, horizontal laces, and round wheels.\}
    } & 
    \fontsize{8pt}{10pt}\selectfont{Q: Hình vẽ số 1 là gì? Nó có đặc điểm gì?
    
    A: Hình vẽ số 1 là một chiếc chìa khóa. Nó có phần đầu tròn, phần thân thẳng và có rãnh ở cuối. Rãnh ở cuối có thể là để khóa hoặc mở một cái gì đó.
    
    Q: Hình vẽ số 2 miêu tả khuôn mặt của ai? Nó biểu hiện cảm xúc gì?
    
    A: Hình vẽ số 2 là một khuôn mặt người. Nó có mái tóc rối, mắt to, mũi ngắn và miệng cười rộng. Nụ cười rộng thể hiện sự vui vẻ, hạnh phúc hoặc chế giễu.
    
    Q: Hình vẽ số 3 là con vật gì? Nó đang làm gì?
    
    A: Hình vẽ số 3 là một con dê. Nó đang đứng yên, đầu nghiêng về phía bên phải. Có thể con dê đang ngửi mùi gì đó hoặc đang nhìn xung quanh.
    
    \hfill \break
    \{Q: What is drawing number 1? What are its features?

A: Drawing number 1 is a key. It has a round head, a straight shaft, and a groove at the end. The groove at the end might be for locking or unlocking something.

Q: Whose face is depicted in drawing number 2? What emotion does it express?

A: Drawing number 2 is a human face. It has messy hair, large eyes, a short nose, and a wide smiling mouth. The wide smile expresses happiness, joy, or possibly mockery.

Q: What animal is drawing number 3? What is it doing?

A: Drawing number 3 is a goat. It is standing still with its head tilted to the right. The goat might be sniffing something or looking around.\}
    } \\
    \end{longtable}

\subsubsection{Handwriting Recognition}
Recognizing handwritten text, especially in diverse handwriting styles, is a challenging task. To address this, we included:
\begin{itemize}
    \item \textbf{Viet-Handwriting-VQA}: Inspired by the Cinnamon AI Challenge - Handwriting Address and UIT-HWDB \cite{nguyen2022uit}, this dataset focuses on improving the model's performance in recognizing handwritten Vietnamese text. 
    
    \item \textbf{Viet-Wiki-Handwriting}: We created a synthesized dataset using various handwriting fonts combined with Wikipedia content, further enriching the training data and improving the model's generalization to different handwriting styles.
\end{itemize}

\begin{longtable}{|m{6.5cm}|m{7.5cm}|}
    \caption{Samples in the Vietnamese handwriting datasets.}
    \label{tab:Viet-Handwriting-VQA-samples} \\
    \hline
    \textbf{Image} & \textbf{Label} \\
    \hline
    \endfirsthead

    \hline
    \textbf{Image} & \textbf{Label} \\
    \hline
    \endhead

    \hline
    \endfoot

    \hline
    \endlastfoot
    \includegraphics[width=6.5cm]{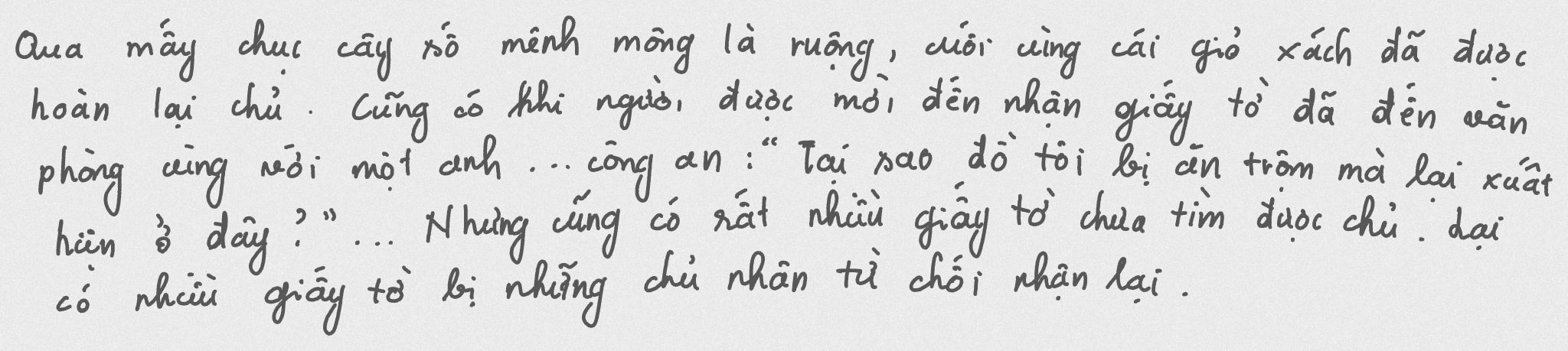} & 
    \fontsize{8.0pt}{10pt}\selectfont{Qua mấy chục cây số mênh mông là ruộng, cuối cùng cái giỏ xách đã được hoàn lại chủ. Cũng có khi người được mời đến nhận giấy tờ đã đến văn phòng cùng với một anh... công an : " Tại sao đồ tôi bị ăn trộm mà lại xuất hiện ở đây? "... Nhưng cũng có rất nhiều giấy tờ chưa tìm được chủ. Lại có nhiều giấy tờ bị những chủ nhân từ chối nhận lại..
    
    \hfill \break
    \{After traveling several dozen kilometers through vast fields, the bag was finally returned to its owner. Sometimes, the person invited to retrieve the documents arrives at the office accompanied by a police officer: "Why did my stolen belongings end up here?"... However, there are still many documents that haven't been claimed by their owners. Additionally, there are documents that the owners have refused to take back.\}
    } \\
    \hline
    \includegraphics[width=6.5cm]{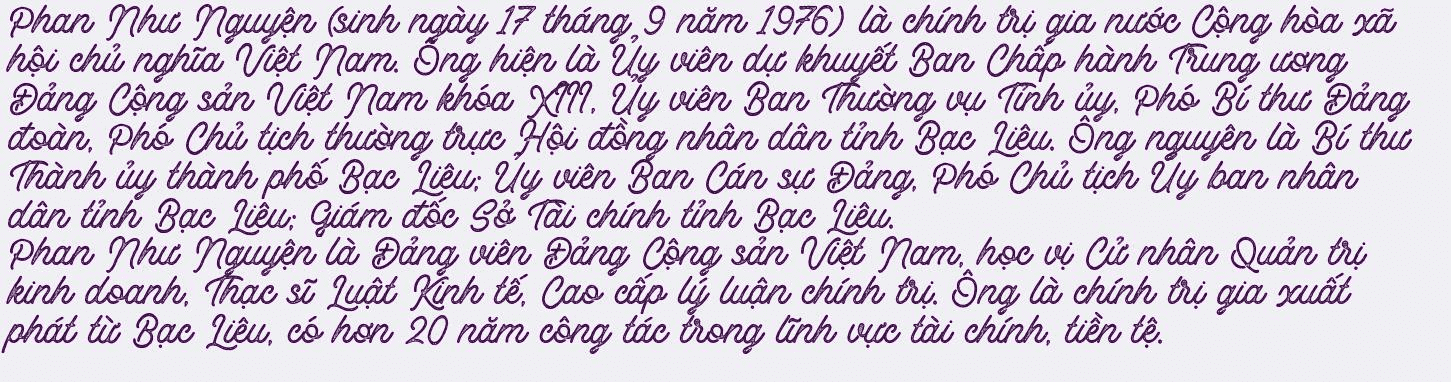} & 
    \fontsize{8.0pt}{10pt}\selectfont{Phan Như Nguyện (sinh ngày 17 tháng 9 năm 1976) là chính trị gia nước Cộng hòa xã hội chủ nghĩa Việt Nam. Ông hiện là Ủy viên dự khuyết Ban Chấp hành Trung ương Đảng Cộng sản Việt Nam khóa XIII, Ủy viên Ban Thường vụ Tỉnh ủy, Phó Bí thư Đảng đoàn, Phó Chủ tịch thường trực Hội đồng nhân dân tỉnh Bạc Liêu. Ông nguyên là Bí thư Thành ủy thành phố Bạc Liêu; Ủy viên Ban Cán sự Đảng, Phó Chủ tịch Ủy ban nhân dân tỉnh Bạc Liêu; Giám đốc Sở Tài chính tỉnh Bạc Liêu. Phan Như Nguyện là Đảng viên Đảng Cộng sản Việt Nam, học vị Cử nhân Quản trị kinh doanh, Thạc sĩ Luật Kinh tế, Cao cấp lý luận chính trị. Ông là chính trị gia xuất phát từ Bạc Liêu, có hơn 20 năm công tác trong lĩnh vực tài chính, tiền tệ.
    
    \hfill \break
    \{Phan Như Nguyện (born September 17, 1976) is a politician of the Socialist Republic of Vietnam. He is currently an Alternate Member of the 13th Central Committee of the Communist Party of Vietnam, a Member of the Standing Committee of the Provincial Party Committee, Deputy Secretary of the Party Committee, and Permanent Vice Chairman of the People's Council of Bạc Liêu Province. He previously served as the Secretary of the Bạc Liêu City Party Committee, Member of the Party Committee of the Provincial People's Committee, Vice Chairman of the People's Committee of Bạc Liêu Province, and Director of the Department of Finance of Bạc Liêu Province. Phan Như Nguyện is a member of the Communist Party of Vietnam, holding a Bachelor's degree in Business Administration, a Master's degree in Economic Law, and an Advanced Degree in Political Theory. He is a politician originating from Bạc Liêu, with over 20 years of experience in finance and monetary sectors.\}
    } \\
    \hline
\end{longtable}

\vfil \break
\subsubsection{Information Extraction}
Information extraction from specific document types is crucial for many applications. For this purpose, we included:
\begin{itemize}
    \item \textbf{Viet-Receipt-VQA}: Based on the MC-OCR 2021 dataset \cite{vu2021mcocr}, this dataset is tailored for extracting information from Vietnamese receipts, such as itemized lists, prices, and dates.
    \item \textbf{Viet-Menu-Gemini-VQA}: Derived from the Quy Nhon AI Hackathon 2022, this dataset focuses on extracting structured information from Vietnamese menus, which involves recognizing dish names, prices, and nutritional information.
\end{itemize}

\begin{longtable}{|m{4cm}|m{3.5cm}|m{4.5cm}|m{4cm}|}
    \caption{Samples in the Viet-Receipt-VQA and Viet-Menu-Gemini-VQA.}
    \label{tab:Viet-information-extraction-samples} \\
    
    \hline
    \textbf{Image} & \textbf{Description} & \textbf{Extractions} & \textbf{Conversations} \\
    \hline
    \endfirsthead

    \hline
    \textbf{Image} & \textbf{Description} & \textbf{Extractions} & \textbf{Conversations} \\
    \hline
    \endhead

    \hline
    \endfoot

    \hline
    \endlastfoot
    \includegraphics[width=4cm]{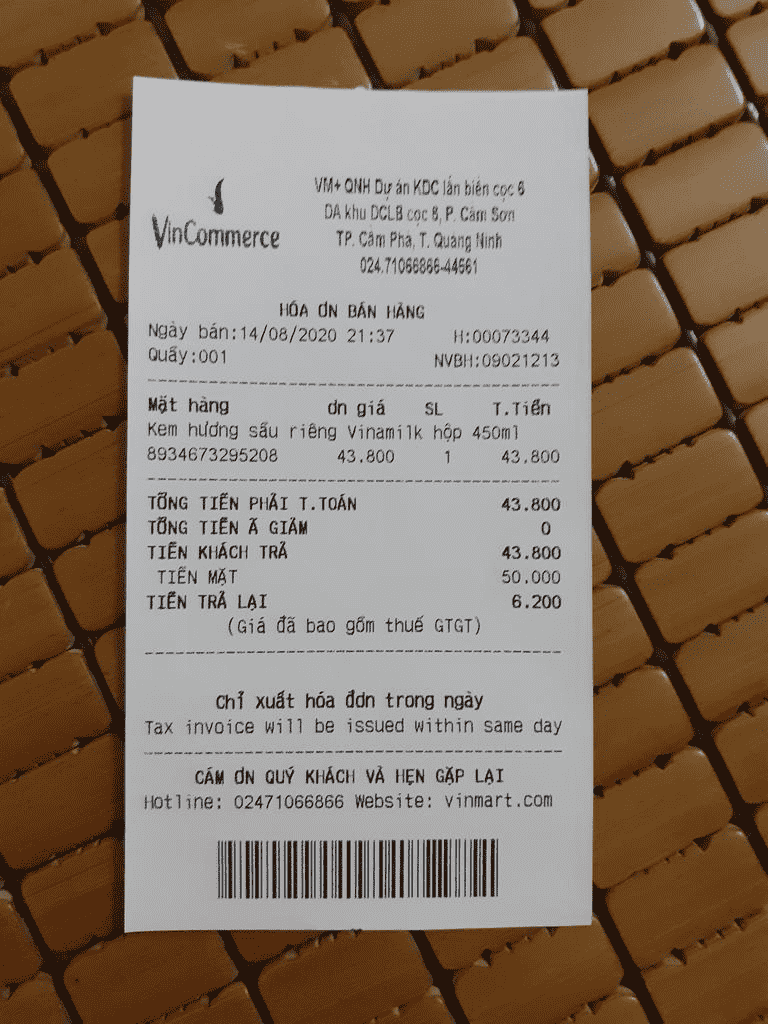} & 
    \fontsize{8.pt}{10pt}\selectfont{Hóa đơn bán hàng của Vincommerce, ngày 14/08/2020, 21:37, mã hóa đơn H:00073344, NVBH:09021213, gồm 1 mặt hàng là Kem hướng sữa riêng Vinamilk hộp 450ml với đơn giá 43.800, tổng tiền phải trả 43.800, tiền khách trả 50.000, tiền trả lại 6.200.

    \hfill \break
    \{Vincommerce Sales Invoice, dated 14/08/2020, 21:37, Invoice Code: H:00073344, Salesperson ID: 09021213. The invoice includes 1 item: Vinamilk Flavored Cream 450ml box, priced at 43,800 VND. The total amount due is 43,800 VND. The customer paid 50,000 VND, and the change given back was 6,200 VND.\}
    } & 
    \fontsize{8.pt}{10pt}\selectfont{\{'Tên cửa hàng': 'Vincommerce', 'Địa chỉ': 'DA khu DCLB cóc 8, P. Cẩm Sơn, TP. Cẩm Phả, T. Quảng Ninh', 'Số điện thoại': '024.71066866-44561', 'Ngày bán': '14\/08\/2020', 'Giờ bán': '21\:37', 'Số quầy': '001', 'Mã hóa đơn': 'H:00073344', 'NVBH': '09021213', 'Mặt hàng': 'Kem hướng sữa riêng Vinamilk hộp 450ml', 'Đơn giá': '43.800', 'Số lượng': '1', 'Tổng tiền': '43.800', 'Tổng tiền à giảm': '0', 'Tiền khách trả': '43.800', 'Tiền mặt': '50.000', 'Tiền trả lại': '6.200'\}
    
    \hfill \break
    \{'Store Name': 'Vincommerce', 'Address': 'DA khu DCLB cóc 8, P. Cẩm Sơn, TP. Cẩm Phả, T. Quảng Ninh', 'Phone Number': '024.71066866-44561', 'Sale Date': '14/08/2020', 'Sale Time': '21:37', 'Counter Number': '001', 'Invoice Code': 'H:00073344', 'Sales Staff': '09021213', 'Item': 'Kem hướng sữa riêng Vinamilk hộp 450ml', 'Unit Price': '43,800', 'Quantity': '1', 'Total Amount': '43,800', 'Total Discounted Amount': '0', 'Amount Paid by Customer': '43,800', 'Cash': '50,000', 'Change Given': '6,200'\}
    } & 
     \fontsize{8.pt}{10pt}\selectfont{Q: Hóa đơn được xuất tại cửa hàng nào?
     
     A: Hóa đơn được xuất tại cửa hàng Vincommerce.
     
     Q: Địa chỉ của cửa hàng là gì?
     
     A: Địa chỉ của cửa hàng là "DA khu DCLB cóc 8, P. Cẩm Sơn, TP. Cẩm Phả, T. Quảng Ninh".
     
     Q: Hóa đơn được xuất vào ngày nào?
     
     A: Hóa đơn được xuất vào ngày 14/08/2020.
     
     Q: Tổng số tiền phải thanh toán là bao nhiêu?
     
     A: Tổng số tiền phải thanh toán là 43.800.
     
     Q: Khách hàng đã thanh toán bằng cách nào?
     
     A: Khách hàng đã thanh toán bằng tiền mặt.

     \hfill \break
     \{Q: At which store was the invoice issued?

A: The invoice was issued at the Vincommerce store.

Q: What is the store's address?

A: The store's address is "DA khu DCLB cóc 8, P. Cẩm Sơn, TP. Cẩm Phả, T. Quảng Ninh".

Q: On what date was the invoice issued?

A: The invoice was issued on 14/08/2020.

Q: What is the total amount to be paid?

A: The total amount to be paid is 43,800.

Q: How did the customer pay?

A: The customer paid in cash.\}
     } \\
    \hline
    
    \includegraphics[width=4cm]{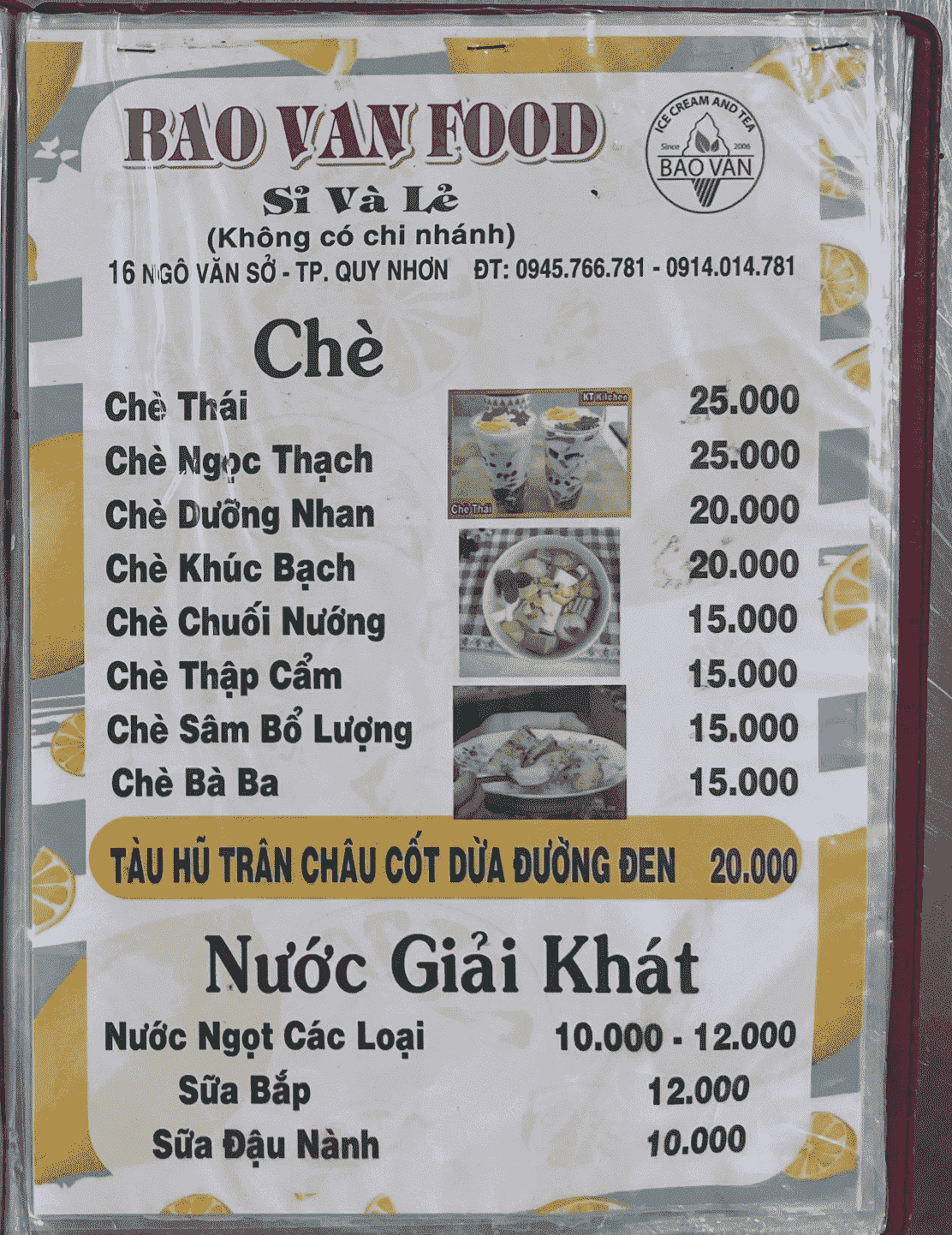} & 
    \fontsize{8.0pt}{10pt}\selectfont{BAO VAN FOOD Si và Lẻ (Không có chi nhánh) 16 NGÕ VÂN SƠ - TP. QUY NHƠN ĐT: 0945.766.781 - 0914.014.781 Chè Chè Thái Chè Ngoc Thach Chè Đường Nhan Chè Khúc Bạch Chè Chuối Nướng Chè Thập Cẩm Chè Sâm Bố Luợng Chè Bá Ba TÂU HỦ TRÂN CHÂU CỐT DỪA ĐƯỜNG ĐEN Nước Giải Khát Nước Ngọt Các Loại Sữa Bắp Sữa Đậu Nành} & 
    \fontsize{8.0pt}{10pt}\selectfont{\{'Tên quán': 'BAO VAN FOOD', 'Loại hình kinh doanh': 'Si và Lẻ', 'Địa chỉ': '16 NGÕ VÂN SƠ - TP. QUY NHƠN', 'Số điện thoại': '0945.766.781 - 0914.014.781', 'Danh sách món ăn': ['Chè Thái', 'Chè Ngoc Thach', 'Chè Đường Nhan', 'Chè Khúc Bạch', 'Chè Chuối Nướng', 'Chè Thập Cẩm', 'Chè Sâm Bố Luợng', 'Chè Bá Ba', 'TÂU HỦ TRÂN CHÂU CỐT DỪA ĐƯỜNG ĐEN', 'Nước Ngọt Các Loại', 'Sữa Bắp', 'Sữa Đậu Nành'], 'Giá tiền': \{'Chè Thái': '25.000', 'Chè Ngoc Thach': '25.000', 'Chè Đường Nhan': '20.000', 'Chè Khúc Bạch': '20.000', 'Chè Chuối Nướng': '15.000', 'Chè Thập Cẩm': '15.000', 'Chè Sâm Bố Luợng': '15.000', 'Chè Bá Ba': '15.000', 'TÂU HỦ TRÂN CHÂU CỐT DỪA ĐƯỜNG ĐEN': '20.000', 'Nước Ngọt Các Loại': '10.000 - 12.000', 'Sữa Bắp': '12.000', 'Sữa Đậu Nành': '10.000'\}\}

    \hfill \break
    \{’Restaurant Name’: ’BAO VAN FOOD’, ’Business Type’: ’Wholesale and Retail’, ’Address’: ’16 NGÕ VÂN SƠ - QUY NHON CITY’, ’Phone Numbers’: ’0945.766.781 - 0914.014.781’, ’Menu Items’: [’Chè Thái’, ’Chè Ngọc Thạch’, ’Chè Đường Nhan’, ’Chè Khúc Bạch’, ’Chè Chuối Nướng’, ’Chè Thập Cẩm’, ’Chè Sâm Bố Lượng’, ’Chè Bá Ba’, ’TÂU HỦ TRÂN CHÂU CỐT DỪA ĐƯỜNG ĐEN’, ’Nước Ngọt Các Loại’, ’Sữa Bắp’, ’Sữa Đậu Nành’], ’Prices’: \{’Chè Thái’: ’25.000’, ’Chè Ngọc Thạch’: ’25.000’, ’Chè Đường Nhan’: ’20.000’, ’Chè Khúc Bạch’: ’20.000’, ’Chè Chuối Nướng’: ’15.000’, ’Chè Thập Cẩm’: ’15.000’, ’Chè Sâm Bố Lượng’: ’15.000’, ’Chè Bá Ba’: ’15.000’, ’TÂU HỦ TRÂN CHÂU CỐT DỪA ĐƯỜNG ĐEN’: ’20.000’, ’Nước Ngọt Các Loại’: ’10.000 - 12.000’, ’Sữa Bắp’: ’12.000’, ’Sữa Đậu Nành’: ’10.000’\}\}
    
    } & 
    \fontsize{8.0pt}{10pt}\selectfont{Q: Tên quán ăn là gì?
    
    A: Tên quán ăn là BAO VAN FOOD.
    
    Q: Quán có bán những loại chè nào?
    
    A: Quán bán các loại chè như: Chè Thái, Chè Ngoc Thach, Chè Đường Nhan, Chè Khúc Bạch, Chè Chuối Nướng, Chè Thập Cẩm, Chè Sâm Bố Luợng, Chè Bá Ba.
    
    Q: Giá của một ly chè Thái là bao nhiêu?
    
    A: Giá của một ly chè Thái là 25.000 đồng.
    
    Q: Quán có bán nước giải khát gì?
    
    A: Quán bán nước ngọt các loại, sữa bắp và sữa đậu nành.
    
    Q: Số điện thoại của quán là gì?
    
    A: Số điện thoại của quán là 0945.766.781 - 0914.014.781.

    \hfill \break
    \{Q: What is the restaurant's name?

A: The restaurant's name is BAO VAN FOOD.

Q: What types of sweet soups does the restaurant sell?

A: The restaurant sells sweet soups such as "Chè Thái, Chè Ngoc Thach, Chè Đường Nhan, Chè Khúc Bạch, Chè Chuối Nướng, Chè Thập Cẩm, Chè Sâm Bố Luợng, Chè Bá Ba".

Q: How much is a cup of Thai Sweet Soup?

A: A cup of Thai Sweet Soup costs 25,000 VND.

Q: What kinds of beverages does the restaurant sell?

A: The restaurant sells various soft drinks, corn milk, and soy milk.

Q: What is the restaurant's phone number?

A: The restaurant's phone numbers are 0945.766.781 - 0914.014.781.\}
    }\\
    \hline
\end{longtable}

\section{Experiments}
\subsection{Implementation Details}
Vintern-1B employed a dynamic high-resolution approach, where images were divided into 448×448 pixel tiles. The number of tiles could vary, reaching up to 12 based on the aspect ratio and resolution during training. In the testing phase, the model could handle as many as 12 tiles.


We applied full-parameter fine-tuning to the Vision Encoder and MLP Projector, and used LoRA \cite{hu2021loralowrankadaptationlarge} for the LLM. We leveraged the pre-trained InternVL2-1B model for visual instruction tuning to fully utilize the MLLM’s capabilities across various multimodal tasks. For next-token prediction, we used cross-entropy loss, consistent with the approach during the pre-training stage.

The initial version of Vintern-1B-v1 was fine-tuned from InternVL2-1B using the Viet-Doc-VQA, Viet-Doc-VQA-II OCR and Viet-OCR-VQA datasets for 1 epoch. In the next phase, Vintern-1B-v2, was further fine-tuned by continuing from v1, using the same Viet-Doc-VQA, Viet-Doc-VQA-II OCR, and Viet-OCR-VQA datasets along with all remaining datasets in \ref{subset:datasets_desc} for an additional 1 epoch. Both versions were trained on 4 GPUs (Nvidia RTX-3090) with a global batch size of 128, a learning rate of 4e-5, and a context length of 4096. Our approach utilizes a context length of 4096, and we follow the response formatting prompts outlined in LLaVA 1.5 \cite{liu2023improvedllava}.

\subsection{Benchmark}




Our evaluation metric is inspired by Lavy \cite{tran2024lavy}, utilizing the MLLM-as-a-Judge approach to verify the accuracy of generated responses for question-answer pairs. Specifically, we employ GPT-4o to assess the quality of answers on two datasets: OpenViVQA \cite{nguyen2023openvivqa} and ViTextVQA \cite{van2024vitextvqa}. We evaluate the zero-shot VQA performance of models on the OpenViVQA-dev \cite{nguyen2023openvivqa} dataset, which comprises 3,505 samples. This dataset presents a challenge for models, requiring them to understand the relationships between Vietnamese images and natural language.

Additionally, we use the ViTextVQA-dev dataset, to evaluate VQA models' capabilities in handling OCR-related tasks, particularly in the context of the Vietnamese language. This dataset primarily focuses on extracting and interpreting information from text and scene text appearing in images.

The evaluation process involves inputting images, questions, labels, and predicted answers into GPT-4o, which then assigns a quality score ranging from 0 to 10 based on the accuracy of the answers. The corresponding results are summarized in Table \ref{table:gpt4o-score}.

Furthermore, we benchmark Vintern-1B-v2 using the VLSP 2023 private test, reporting the F1 and BLEU scores, as shown in Table \ref{tab:vlsp2023benchmarks}. It is important to note that the F1 and BLEU scores are relatively may not be as high because the Vintern-1B model tends to generate longer responses than the reference answers, which adversely impacts the scoring metrics. Some qualitative test cases
are showed in Table \ref{tab:qualitities-test-cases}.



\begin{longtable}[c]{|l|c|c|}
\caption{Zero-shot VQA on OpenViVQA-dev and ViTextVQA-dev. Models’ output accuracy are evaluated by GPT-4o}
\label{table:gpt4o-score} \\
\hline
\textbf{Model} & \multicolumn{2}{c|}{\textbf{GPT-4o-score}} \\
\cline{2-3}
& \textbf{OpenVivQA-dev} & \textbf{ViTextVQA-dev} \\
\hline
\endfirsthead

\hline
\textbf{Model} & \multicolumn{2}{c|}{\textbf{GPT-4o-score}} \\
\cline{2-3}
& \textbf{OpenVivQA-dev} & \textbf{ViTextVQA-dev} \\
\hline
\endhead

\hline
\endfoot

\hline
\endlastfoot

Vintern-1B-v1 & $7.1 / 10$ & $7.6 / 10$ \\
\hline
Vintern-1B-v2 & $\mathbf{7.7 / 10}$ & $\mathbf{7.7 / 10}$ \\

\end{longtable}

    


\vfil \break
\begin{longtable}{|l|c|c|}
\caption{VLSP 2023 Benchmarks}
\label{tab:vlsp2023benchmarks} \\
\hline
\textbf{Team Name} & \textbf{F1} & \textbf{avg. BLEU} \\
\hline
\endfirsthead

\hline
\textbf{Team Name} & \textbf{F1} & \textbf{avg. BLEU} \\
\hline
\endhead

\hline
\endfoot

\hline
\endlastfoot

ICNLP & 3.6384 (1) & 0.4663 (4) \\
\textbf{Vintern-1B-v2} & 3.4616 (2) & 0.4422 (7) \\
linh & 3.4293 (3) & 0.4609 (5) \\
DS@ViVRC & 3.4121 (4) & 0.4457 (6) \\
DS@UIT Dynasty & 3.3172 (5) & 0.4742 (3) \\
NTQ Solution & 3.2926 (6) & 0.4876 (1) \\
I, Me \& Myself & 3.2396 (7) & 0.4795 (2) \\
AVQA\_AIO & 2.9018 (8) & 0.4380 (8) \\
\textbf{Vintern-1B-v1} & 2.7256 (9) & 0.3082 (10) \\
NguyenLe & 2.7053 (10) & 0.3997 (9) \\
nowj2 & 1.6808 (11) & 0.2412 (11) \\

\end{longtable}

\begin{longtable}{|m{6cm}|m{3.5cm}|m{6.0cm}|}
    \caption{Qualitative test cases of Vintern-1B-v2.}
    \label{tab:qualitities-test-cases} \\
    \hline
    \textbf{Image} & \textbf{Question}  & \textbf{Model Answer}\\
    \hline
    \endfirsthead

    \hline
    \textbf{Image} & \textbf{Question}  & \textbf{Model Answer} \\
    \hline
    \endhead

    \hline
    \endfoot

    \hline
    \endlastfoot

\includegraphics[width=6cm]{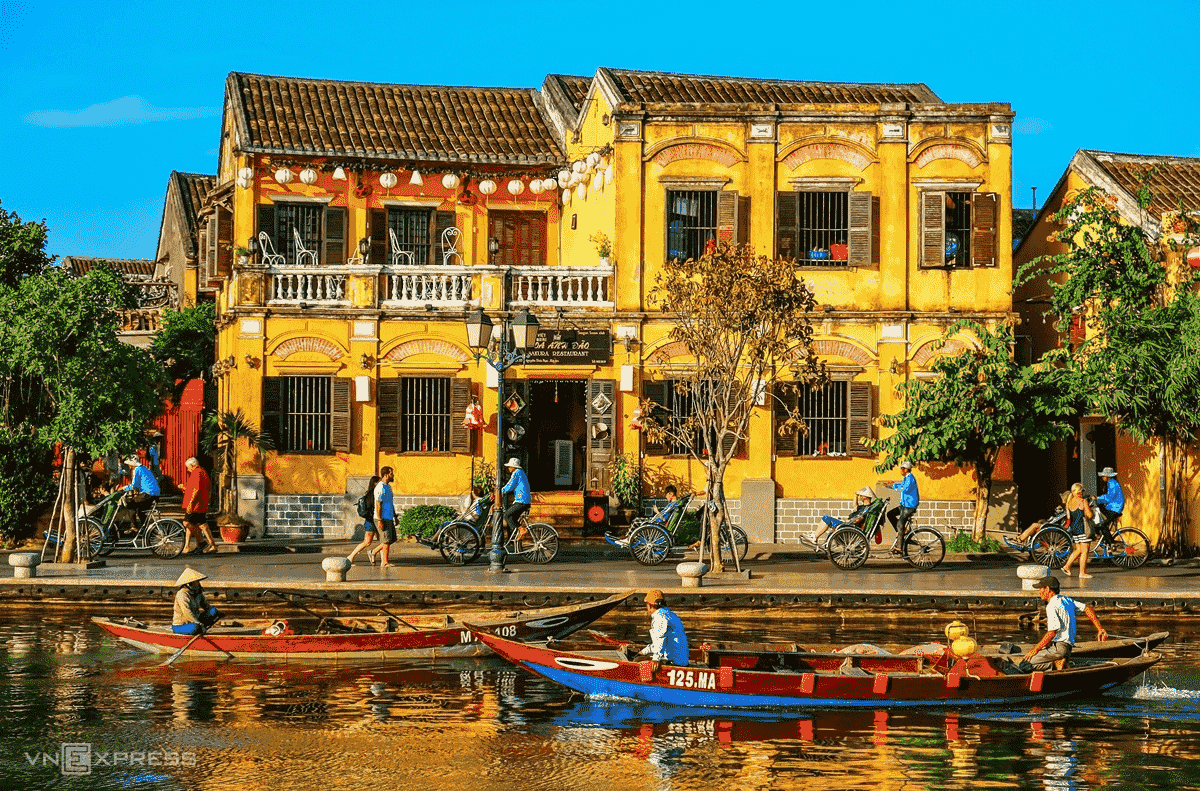} & 
    \fontsize{8.0pt}{10pt}\selectfont{Mô tả hình ảnh một cách chi tiết.}
    
    \hfill \break
    \fontsize{8.0pt}{10pt}\selectfont\{Detailed description of the image.\}
    & 
    \fontsize{8.0pt}{10pt}\selectfont{Bức ảnh chụp một con sông ở Việt Nam với những ngôi nhà cổ kính được sơn màu vàng nhạt. Phía trước là một con thuyền nhỏ màu đỏ và một chiếc thuyền lớn màu xanh dương.
Trên thuyền có một người đàn ông đang chèo thuyền. Hai bên bờ sông là những hàng cây xanh mát. Bầu trời trong xanh và có một vài đám mây trắng.
Ở góc dưới bên trái của bức ảnh có dòng chữ "VNEXPRESS".

\hfill \break
\{The photo captures a river in Vietnam with ancient houses painted in light yellow. In the foreground, there is a small red boat and a larger blue boat. On the boat, there is a man rowing. Along both sides of the river are rows of green trees. The sky is clear blue with a few white clouds. In the bottom left corner of the photo, there is the text "VNEXPRESS."\}

} \\
    \hline

    \includegraphics[width=6cm]{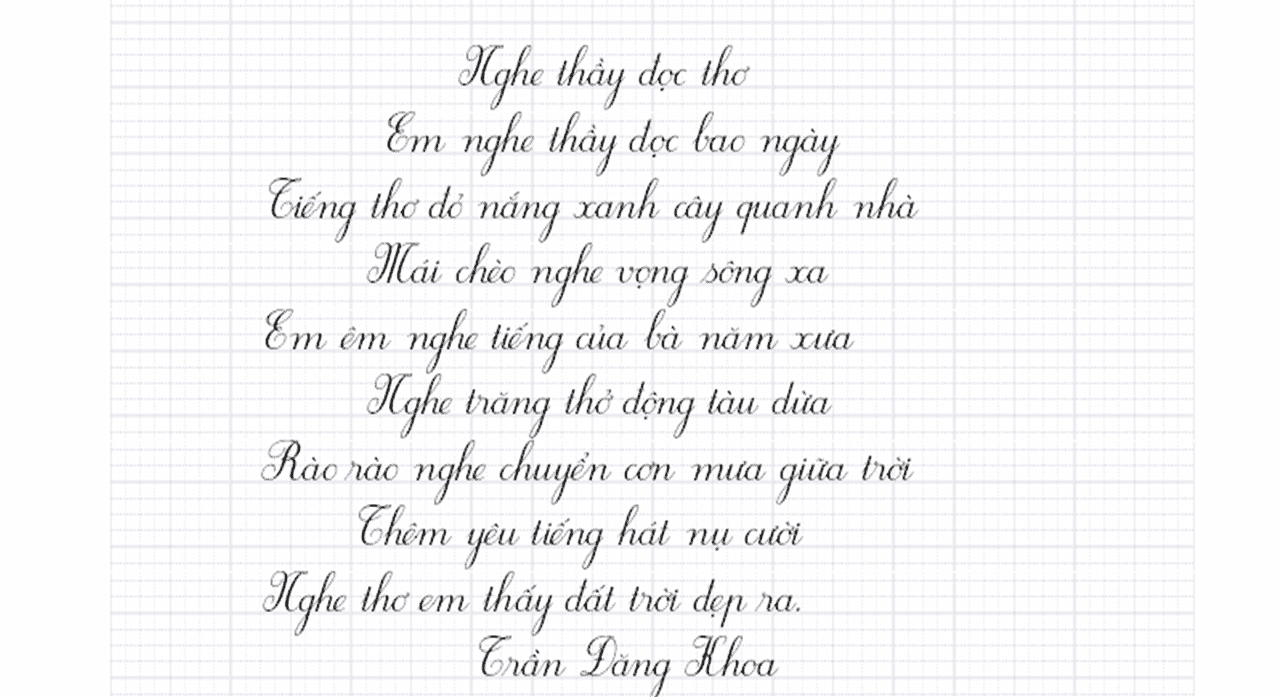} & 
    \fontsize{8.0pt}{10pt}\selectfont{Mô tả hình ảnh một cách chi tiết.}

    \hfill \break
    \fontsize{8.0pt}{10pt}\selectfont{\{Detailed description of the image.\}}
    & 
    \fontsize{8.0pt}{10pt}\selectfont{Bức ảnh là một trang giấy kẻ ô vuông với dòng chữ viết tay bằng mực đen. Bố cục đơn giản, chỉ có nội dung văn bản được viết theo chiều dọc từ trên xuống dưới. Các dòng chữ được sắp xếp theo thứ tự từ trên xuống dưới như sau:

- Nghe thầy đọc thơ

- Em nghe thầy đọc bao ngày

- Tiếng thơ đỏ nắng xanh cây quanh nhà

- Mái chèo nghe vọng sông xa

- Em êm nghe tiếng của bà năm xưa

- Nghe trăng thở động tàu dừa

- Rào rào nghe chuyển con mưa giữa trời

- Thêm yêu tiếng hát nụ cười

- Nghe thơ em thấy đất trời đẹp ra.

- Trần Đăng Khoa

\hfill \break
\{The image is of a grid-lined page with handwritten text in black ink. The layout is simple, containing only the text, which is written vertically from top to bottom. The lines of text are arranged in the following order:

- Nghe thầy đọc thơ

- Em nghe thầy đọc bao ngày

- Tiếng thơ đỏ nắng xanh cây quanh nhà

- Mái chèo nghe vọng sông xa

- Em êm nghe tiếng của bà năm xưa

- Nghe trăng thở động tàu dừa

- Rào rào nghe chuyển con mưa giữa trời

- Thêm yêu tiếng hát nụ cười

- Nghe thơ em thấy đất trời đẹp ra.

- Trần Đăng Khoa\}

} \\

    \hline
    \includegraphics[width=6cm]{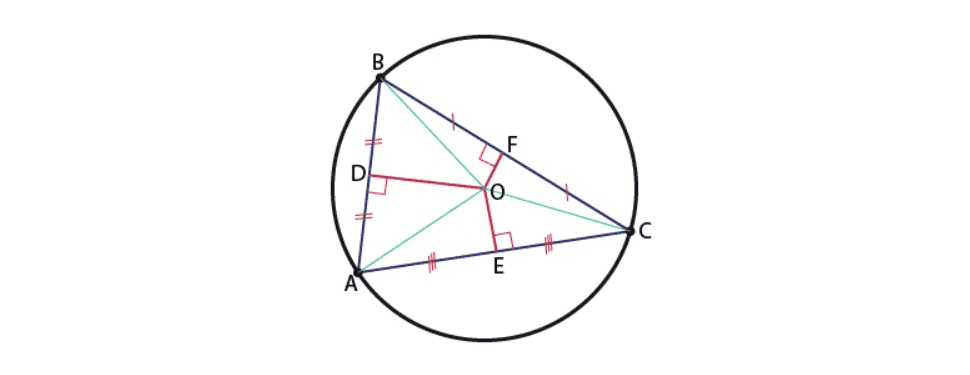} & 
    \fontsize{8.0pt}{10pt}\selectfont{Tam giác ABC là gì của đường tròn tâm O ?}

    \hfill \break
    \fontsize{8.0pt}{10pt}\selectfont{ \{What is triangle ABC with respect to the circle centered at O?\} }
    & 
    \fontsize{8.0pt}{10pt}\selectfont{Tam giác ABC là tam giác nội tiếp của đường tròn tâm O.
    
    \hfill \break
    \fontsize{8.0pt}{10pt}\selectfont{\{Triangle ABC is the inscribed triangle of the circle centered at O.\}}
    } \\
    \hline

     \includegraphics[width=6cm]{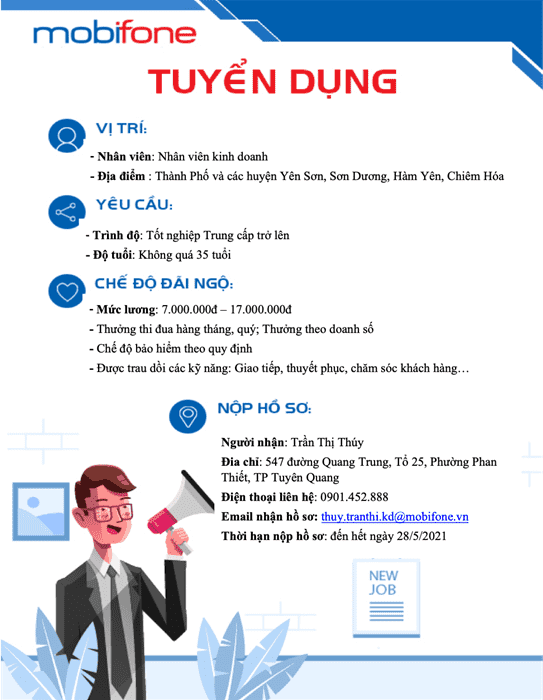} & 
    \fontsize{8.0pt}{10pt}\selectfont{Mô tả hình ảnh một cách chi tiết.} 

    \hfill \break
    \fontsize{8.0pt}{10pt}\selectfont{\{Detailed description of the image.\}}
    & 
    \fontsize{8.0pt}{10pt}\selectfont{Bức ảnh là một thông báo tuyển dụng của công ty Mobifone. Bố cục được chia thành 3 phần chính:

- Phần đầu tiên là tiêu đề "TUYỂN DỤNG" được in chữ màu đỏ đậm, cỡ chữ lớn, nằm ở vị trí trung tâm phía trên cùng của bức ảnh.

- Phần thứ hai là phần giới thiệu về vị trí và yêu cầu của ứng viên. Phần này được trình bày theo dạng liệt kê các mục:
    - Vị trí: Nhân viên kinh doanh
    - Địa điểm: Thành phố và các huyện Yên Sơn, Sơn Dương, Hàm Yên, Chiêm Hóa
    - Yêu cầu: Trình độ: Tốt nghiệp Trung cấp trở lên, Độ tuổi: Không quá 35 tuổi

- Phần thứ ba là phần giới thiệu về chế độ đãi ngộ. Phần này được trình bày theo dạng liệt kê các mục:
    - Chế độ đãi ngộ:
        - Mức lương: 7.000.000đ - 17.000.000đ
        - Thưởng thi đua hàng tháng, quý: Thưởng theo doanh số
        - Chế độ bảo hiểm theo quy định
        - Được trau dồi các kỹ năng: Giao tiếp, thuyết phục, chăm sóc khách hàng...

Phần cuối cùng là phần "NỘP HỒ SƠ" với thông tin liên lạc của người nộp hồ sơ. Phần này được trình bày theo dạng liệt kê các mục:
    - Người nhận: Trần Thị Thúy
    - Địa chỉ: 547 đường Quang Trung, Tổ 25, Phường Phan Thiết, TP Tuyên Quang
    - Điện thoại liên hệ: 0901.452.888
    - Email nhận hồ sơ: thuy.tranthi.kd@mobifone.vn
    - Thời hạn nộp hồ sơ: đến hết ngày 28/5/2021
    
    \hfill \break
    \{The picture is a job announcement from Mobifone. The layout is divided into three main sections:

- The first section is the title 'RECRUITMENT,' printed in bold red letters, large font size, centered at the top of the image.

- The second section introduces the position and requirements for the candidates. This section is presented in a bullet point format: - Position: Sales Staff - Location: The city and districts of Yên Sơn, Sơn Dương, Hàm Yên, Chiêm Hóa - Requirements: Education: At least a vocational diploma, Age: Not over 35 years old.

- The third section introduces the benefits package. This section is also presented in a bullet point format: - Benefits: - Salary: 7,000,000 VND - 17,000,000 VND - Monthly and quarterly performance bonuses: Sales-based bonuses - Insurance benefits as per regulations - Opportunities to develop skills such as communication, persuasion, customer service, etc.
The final section is the 'SUBMIT APPLICATION' part with the contact information for the applicant. This section is presented in a bullet point format: - Recipient: Trần Thị Thúy - Address: 547 Quang Trung Street, Group 25, Phan Thiết Ward, Tuyên Quang City - Contact phone number: 0901.452.888 - Email for submission: thuy.tranthi.kd@mobifone.vn - Application deadline: until the end of May 28, 2021.\}
    } \\
    \hline

    \includegraphics[width=6cm]{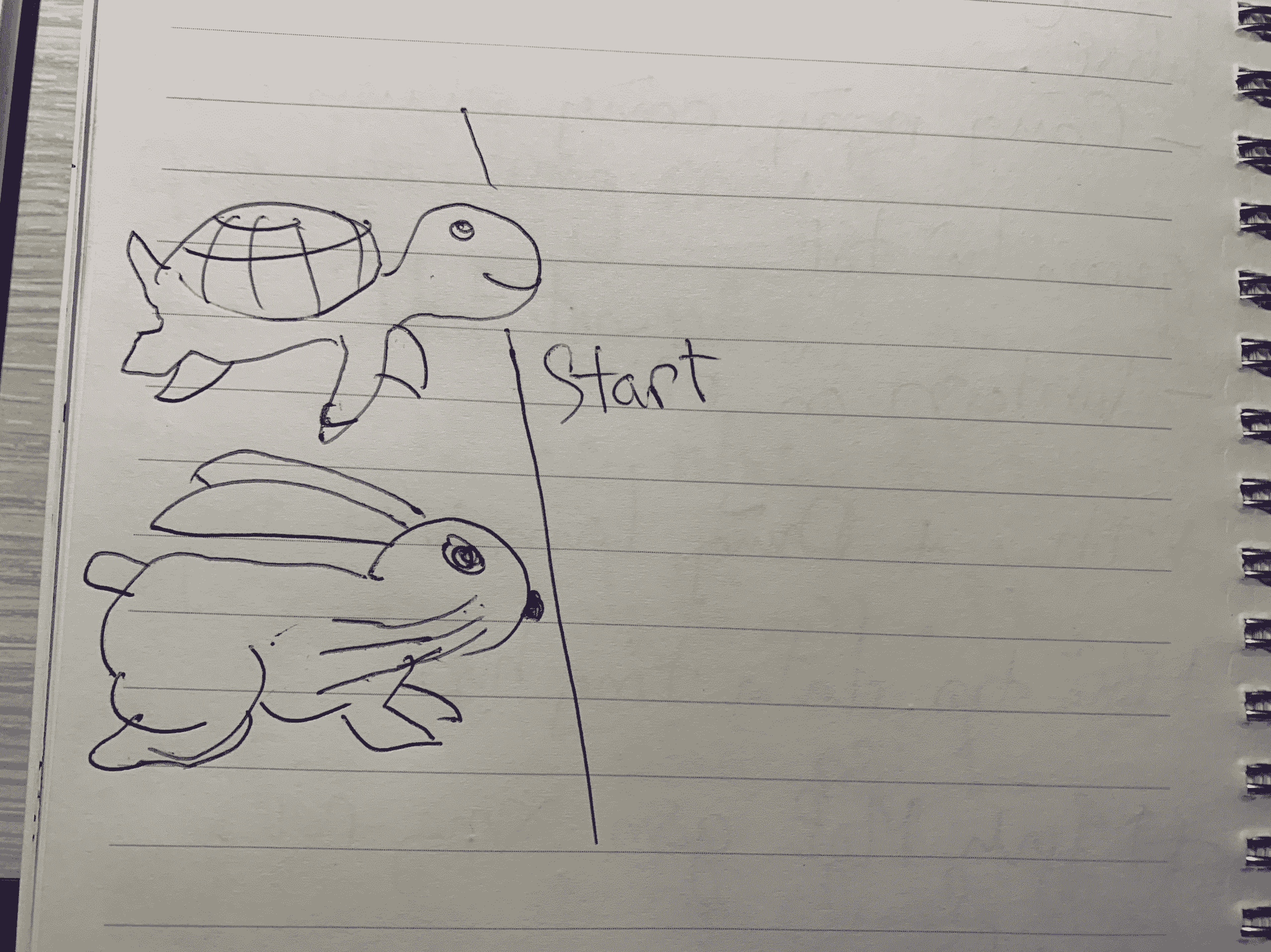} & 
    \fontsize{8.0pt}{10pt}\selectfont{Con nào sẽ thắng ? }
    
    \hfill \break
    \fontsize{8.0pt}{10pt}\selectfont{\{Which one will win?\}}
    & \fontsize{8.0pt}{10pt}\selectfont{Con thỏ sẽ thắng.

    \hfill \break

    \{The rabbit will win.\}
    } \\
    \hline

    \includegraphics[width=6cm]{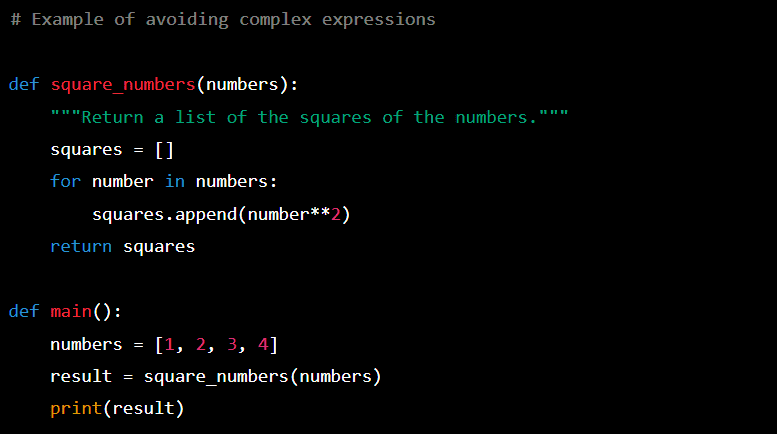} & 
    \fontsize{8.0pt}{10pt}\selectfont{Mô tả chi tiết hình ảnh.}
    
    \hfill \break
    \fontsize{8.0pt}{10pt}\selectfont{\{Detailed description of the image.\}}
    & 
    \fontsize{8.0pt}{10pt}\selectfont{Bức ảnh hiển thị một đoạn mã nguồn được viết bằng ngôn ngữ lập trình Python. Đoạn mã này bao gồm hai phần:
Phần đầu: Nêu tiêu đề "Example of avoiding complex expressions" (Mô tả ví dụ về việc tránh các hàm phức tạp).
Phần chính: Bao gồm hai hàm square\_numbers và main().
Hàm square\_numbers nhận một tham số là danh sách số từ 1 đến 4. Hàm này sử dụng vòng lặp for để duyệt qua mỗi số trong danh sách và thêm nó vào danh sách squares. Sau đó, hàm trả về danh sách squares.
Hàm main() tạo một biến numbers với giá trị [1, 2, 3, 4], sau đó gọi hàm square\_numbers và ghi kết quả vào biến result. Cuối cùng, hàm print(result) sẽ in ra kết quả của hàm main().

\hfill \break
\{The image displays a piece of source code written in Python. The code includes two parts:

Title: "Example of avoiding complex expressions".

Main part: It includes two functions: square\_numbers and main().

The square\_numbers function takes a list of numbers from 1 to 4 as an argument. This function uses a for loop to iterate through each number in the list and adds it to the squares list. Then, the function returns the squares list.

The main() function creates a variable numbers with the value [1, 2, 3, 4], then calls the square\_numbers function and stores the result in the result variable. Finally, the print(result) function prints out the result of the main() function.\}

} \\
    \hline
\end{longtable}

\section{Conclusion}
In conclusion, Vintern-1B marks a pioneering step forward in the development of multimodal models for the Vietnamese language. By integrating state-of-the-art language and visual models with high-quality datasets specific to Vietnamese, Vintern-1B has shown impressive performance across a variety of tasks within the Vietnamese context. It excels particularly in OCR-related tasks and demonstrates significant improvements in Vietnamese-related scene understanding and complex reasoning.

The model’s optimized design ensures that it remains compact enough for deployment on edge devices, which broadens its usability across a wide range of applications. Additionally, its creation and fine-tuning on large-scale multimodal datasets tailored specifically for Vietnamese make it not only effective but also highly relevant for the local context.

Overall, Vintern-1B sets a new benchmark for Vietnamese multimodal models, providing a solid foundation for future research and development in this field. Its open-source availability invites further collaboration and innovation, helping to bridge the gap in resources and capabilities for Vietnamese language technology.
\vfil \break



\bibliographystyle{IEEEtran}
\bibliography{references}

\end{document}